%% file: New_IEEEtran_how-to.tex
\documentclass[lettersize,journal]{IEEEtran}
\usepackage{amsmath,amsfonts}
\usepackage{array}
\usepackage[caption=false,font=normalsize,labelfont=sf,textfont=sf]{subfig}
\usepackage{textcomp}
\usepackage{setspace}    
\usepackage{tabularx}
\makeatletter
\usepackage{url}
\usepackage{verbatim}
\usepackage{graphicx}
\usepackage{tabularx}
\usepackage{booktabs}   
\usepackage{multirow}    
\usepackage{graphicx}    
\usepackage[utf8]{inputenc}
\DeclareUnicodeCharacter{2212}{-}
\usepackage{algorithm}
\usepackage{tabularx}
\usepackage[T1]{fontenc}
\usepackage{amsmath}       

\usepackage{algorithm}     
\usepackage{algpseudocode} 

\usepackage{tikz}
\usepackage{forest}
\usepackage{xcolor}
\usepackage[most]{tcolorbox}

\tcbset{
  badprompt/.style={
    colback=red!3,
    colframe=red!40,
    fonttitle=\bfseries,
    title=#1
  }
}
\usepackage{xr-hyper}
\usepackage{hyperref}



\def\BibTeX{{\rm B\kern-.05em{\sc i\kern-.025em b}\kern-.08em
    T\kern-.1667em\lower.7ex\hbox{E}\kern-.125emX}}
\usepackage{balance}
\begin{document}
\title{A Lightweight LLM Framework for Disaster Humanitarian Information Classification}
\author{Han~Jinzhen,
        Kim~Jisung,
        Yang~Jong~Soo,
        and~Yun~Hong~Sik%
\thanks{This work has been submitted to the IEEE for possible publication. Copyright may be transferred without notice, after which this version may no longer be accessible.}
\thanks{Manuscript received [Date]. This research was supported by the 2023-MOIS36-004 (RS-2023-00248092) of the Technology Development Program on Disaster Restoration Capacity Building and Strengthening funded by the Ministry of Interior and Safety (MOIS, Korea). (Corresponding author: Yun Hong Sik).}%
\thanks{Han Jinzhen, Yang Jong Soo, and Yun Hong Sik are with the Department of Civil, Architectural \& Environment Engineering, Sungkyunkwan University, Suwon, Korea (e-mail: hanjinzhen9@gmail.com).}%
\thanks{Kim Jisung is with the School of Geography, University of Leeds, LS2 9JT, United Kingdom.}%
}

\markboth{Journal of \LaTeX\ Class Files,~Vol.~18, No.~9, September~2020}%
{How to Use the IEEEtran \LaTeX \ Templates}

\maketitle

\begin{abstract}
Timely classification of humanitarian information from social media is critical for effective disaster response. However, deploying large language models (LLMs) for this task faces challenges in resource-constrained emergency settings. This paper develops a lightweight, cost-effective framework for disaster tweet classification using parameter-efficient fine-tuning. We construct a unified experimental corpus by integrating and normalizing the HumAID dataset (76,484 tweets across 19 disaster events) into a dual-task benchmark: humanitarian information categorization and event type identification. Through systematic evaluation of prompting strategies, LoRA fine-tuning, and retrieval-augmented generation (RAG) on Llama 3.1 8B, we demonstrate that: (1) LoRA achieves 79.62\% humanitarian classification accuracy (+37.79\% over zero-shot) while training only $\sim$2\% of parameters; (2) QLoRA enables efficient deployment with 99.4\% of LoRA performance at 50\% memory cost; (3) contrary to common assumptions, RAG strategies degrade fine-tuned model performance due to label noise from retrieved examples. These findings establish a practical, reproducible pipeline for building reliable crisis intelligence systems with limited computational resources.

\end{abstract}


\section{Introduction}
Social networking services (SNS), particularly microblog platforms such as Twitter/X, have become core infrastructures for disaster communication and coordination. Crisis informatics research shows that networked publics can rapidly produce and circulate situational information that complements official channels, reshaping emergency communication and enabling bottom-up sensemaking during disruptive events \cite{palenCitizenCommunicationsCrisis2007,liuOrganizationalDisasterCommunication2021}. Empirical studies further demonstrate that SNS posts often contain actionable content---warnings, damage reports, and requests---with distinctive temporal dynamics and interaction patterns \cite{viewegMicrobloggingTwoNatural2010,mendozaTwitterCrisisCan2010}. Volunteer-driven, self-organizing behaviors also emerge, enabling distributed coordination and information brokerage when formal resources are strained \cite{starbirdVoluntweetersSelforganizingDigital2011}.

Given the high-volume, noisy, and fast-evolving nature of crisis SNS streams, early computational work focused on event detection, filtering, and credibility assessment. Real-time SNS signals can capture sudden-onset hazards such as earthquakes \cite{sakakiEarthquakeShakesTwitter2010}, but credibility varies substantially across posts and must be modeled explicitly \cite{castilloInformationCredibilityTwitter2011}. Geographic perspectives further highlight uneven spatial distributions of crisis information and the value of incorporating location context for interpretation and relevance ranking \cite{crooksEarthquakeTwitterDistributed2013,dealbuquerqueGeographicApproachCombining2015}. Operational systems have also been proposed to convert SNS streams into alerting pipelines, including earthquake alerting frameworks based on social signals \cite{avvenutiEARSEarthquakeAlert2014}.

To support more systematic crisis analytics, supervised NLP for humanitarian information extraction has been advanced through annotated datasets and shared tasks \cite{imranProcessingSocialMedia2015}. These efforts established practical taxonomies (e.g., casualties, infrastructure damage, evacuation, urgent needs), but classic supervised models remain vulnerable to domain shift, evolving vocabulary, and label ambiguity in real-world deployments, motivating renewed interest in foundation models and large language models (LLMs).

Recent LLM-centered efforts extend crisis analytics beyond classification to crisis-aware reasoning, summarization, retrieval-augmented workflows, and question answering. Prior work includes instruction-oriented crisis sensemaking with fine-tuned LLMs \cite{yinCrisisSenseLLMInstructionFineTuned2024}, flood-related decision support resources \cite{colverdFloodBrainFloodDisaster2023}, crisis response management systems \cite{lamsalCReMaCrisisResponse2024}, crisis summarization \cite{seebergerCrisis2SumExploratoryStudy2024}, benchmark-style disaster QA evaluation \cite{rawatDisasterQABenchmarkAssessing2024}, and broader discussions of LLM-assisted crisis management workflows \cite{otalLLMAssistedCrisisManagement2024}. While LLMs offer improved semantic generalization and instruction following, they also introduce new reliability concerns.

Key limitations remain for high-stakes humanitarian classification from SNS. Robustness under severe class imbalance and ambiguous category boundaries is challenging even with strong foundation models, requiring reliability-oriented evaluation \cite{kalluriRobustDisasterAssessment2024}. Moreover, operational settings demand timely identification of actionable information under constraints on latency, cost, and expert labeling, motivating sustainable and efficient pipelines rather than accuracy-centric prototypes \cite{chenSustainableAgileIdentification2025}.

This study develops a lightweight and controllable LLM-based framework for \textbf{dual-task classification} of disaster-related SNS messages: (1) humanitarian information categorization into 10 fine-grained classes (e.g., casualties, infrastructure damage, urgent needs, rescue efforts), and (2) disaster event type identification across 4 categories (earthquake, fire, flood, hurricane). Using the HumAID dataset with 76,484 annotated tweets and Llama 3.1 8B as the base model, we evaluate prompting, parameter-efficient fine-tuning (LoRA), and retrieval-augmented generation (RAG).

Before adaptation, a raw instruction-tuned LLM is inadequate for humanitarian classification: zero-shot prompting achieves 41.83\% accuracy and dynamic few-shot improves to 64.10\%. Event type classification is easier, with zero-shot reaching 62.74\%. This gap suggests that generic instruction following is insufficient to resolve humanitarian label intent in short, noisy, and ambiguous SNS text, especially for overlapping categories such as \texttt{other\_relevant\_information} and \texttt{not\_humanitarian}.

LoRA fine-tuning substantially improves performance to 79.62\% accuracy for humanitarian classification and 98.79\% for event type classification while updating only $\sim$2\% of model parameters, aligning with operational needs for rapid, reproducible adaptation. QLoRA (4-bit quantization) retains 99.4\% of LoRA performance while reducing memory use by 50\%, supporting deployment in resource-constrained settings.

After LoRA adaptation, RAG is not consistently beneficial for fine-grained humanitarian classification. Standard RAG reduces accuracy from 79.62\% to 77.53\%, and adaptive or hybrid variants do not exceed the LoRA baseline. Error analysis indicates a mismatch between semantic similarity and label intent: retrieved examples are often topically similar (same disaster type) but belong to different humanitarian categories (e.g., \emph{damage} vs.\ \emph{casualties}), biasing predictions. This challenges the common assumption that retrieval augmentation universally improves LLM performance.

GPT-4 validation suggests that low accuracy on certain category pairs reflects \textbf{inherent task ambiguity} rather than model limitations. GPT-4 achieves only 21.67\% accuracy on the top 10 confusion pairs, and only 3.33\% on \texttt{other\_relevant\_information} vs.\ \texttt{not\_humanitarian}. This implies that performance ceilings for some categories are constrained by taxonomy definitions, and that category refinement or merging may be more effective than further optimization.

The contributions of this work are summarized as follows:
\begin{itemize}
    \item A systematic empirical study of LLM adaptation strategies for dual-task disaster SNS classification, covering zero-shot, few-shot, LoRA fine-tuning, and RAG, and clarifying what works and under which conditions.
    \item Evidence that LoRA fine-tuning strongly outperforms prompting (+37.79\% absolute for humanitarian classification, +36.05\% for event type), supporting parameter-efficient fine-tuning as the preferred strategy.
    \item An empirical explanation of when and why RAG can \emph{harm} a strong fine-tuned classifier, demonstrated across standard, adaptive, and hybrid RAG strategies.
    \item GPT-4-based evidence that certain low-accuracy categories reflect intrinsic ambiguity in the taxonomy, motivating principled category refinement or merging.
    \item A reproducible, cost-effective pipeline combining LoRA/QLoRA fine-tuning with confusion-aware analysis for operational crisis intelligence systems.
\end{itemize}

\section{Data and Preprocessing}

This research uses HumAID (Human-Annotated Disaster Incidents Data), a large-scale Twitter dataset curated for humanitarian information processing in disaster contexts \cite{alam2021humaid}. HumAID provides human-labeled English tweets sampled from a much larger pool of disaster-related posts across multiple real-world events, together with deep-learning benchmarks and standardized evaluation splits. These properties make HumAID an appropriate testbed for examining fine-grained humanitarian tweet classification under realistic, high-noise crisis-stream conditions.

Figure~\ref{fig:preprocessing} illustrates the data preprocessing pipeline employed in this study. The pipeline transforms the raw HumAID dataset into a unified JSONL format suitable for downstream humanitarian information classification tasks.

\usetikzlibrary{positioning, arrows.meta, shapes.geometric, fit, calc, backgrounds}

\definecolor{rawbg}{HTML}{E3F2FD}
\definecolor{rawborder}{HTML}{1976D2}
\definecolor{rawtitle}{HTML}{1565C0}
\definecolor{procbg}{HTML}{FFF3E0}
\definecolor{procborder}{HTML}{E65100}
\definecolor{outbg}{HTML}{E8F5E9}
\definecolor{outborder}{HTML}{388E3C}
\definecolor{outtitle}{HTML}{2E7D32}
\definecolor{folderbg}{HTML}{FFF9C4}
\definecolor{folderborder}{HTML}{FFA000}
\definecolor{filebg}{HTML}{FFFFFF}
\definecolor{fileborder}{HTML}{9E9E9E}
\definecolor{arrowcolor}{HTML}{E53935}
\definecolor{totalbox}{HTML}{C8E6C9}

\usetikzlibrary{positioning, arrows.meta, backgrounds}

\definecolor{rawbg}{HTML}{E3F2FD}
\definecolor{rawborder}{HTML}{1976D2}
\definecolor{rawtitle}{HTML}{1565C0}
\definecolor{procbg}{HTML}{FFF3E0}
\definecolor{procborder}{HTML}{E65100}
\definecolor{outbg}{HTML}{E8F5E9}
\definecolor{outborder}{HTML}{388E3C}
\definecolor{outtitle}{HTML}{2E7D32}
\definecolor{folderbg}{HTML}{FFF9C4}
\definecolor{folderborder}{HTML}{FFA000}
\definecolor{filebg}{HTML}{FFFFFF}
\definecolor{fileborder}{HTML}{9E9E9E}
\definecolor{arrowcolor}{HTML}{E53935}
\definecolor{totalboxcolor}{HTML}{C8E6C9}

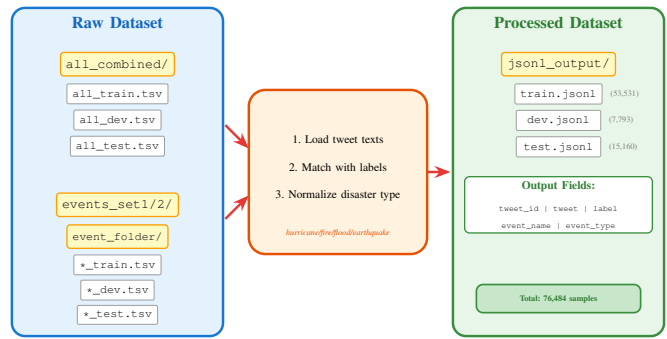
\begin{figure}[!t]
    \centering

    \resizebox{\linewidth}{!}{
        \begin{tikzpicture}[
            folder/.style={
                rectangle, rounded corners=2pt, draw=folderborder, fill=folderbg,
                minimum width=2.2cm, minimum height=0.5cm, 
                font=\footnotesize\ttfamily, align=center, line width=0.8pt
            },
            file/.style={
                rectangle, rounded corners=1pt, draw=fileborder, fill=filebg,
                minimum width=1.8cm, minimum height=0.4cm,
                font=\scriptsize\ttfamily, align=center
            },
            bigarrow/.style={
                -Stealth, line width=1.5pt, color=arrowcolor
            }
        ]
        
        \begin{scope}[on background layer]
            \node[rectangle, rounded corners=8pt, draw=rawborder, fill=rawbg, line width=1.5pt,
                  minimum width=4.5cm, minimum height=7cm] (rawbox) at (1.8, 4.5) {};
        \end{scope}
        \node[font=\normalsize\bfseries, text=rawtitle] at (1.8, 7.7) {Raw Dataset};

        \node[folder] (ac) at (1.8, 6.8) {all\_combined/};
        \node[file, below=0.15cm of ac] (ac1) {all\_train.tsv};
        \node[file, below=0.12cm of ac1] (ac2) {all\_dev.tsv};
        \node[file, below=0.12cm of ac2] (ac3) {all\_test.tsv};

        \node[folder] (es) at (1.8, 3.8) {events\_set1/2/};
        \node[folder, below=0.15cm of es, minimum width=1.8cm, font=\scriptsize\ttfamily] (ef) {event\_folder/};
        \node[file, below=0.12cm of ef, minimum width=1.5cm] (ef1) {*\_train.tsv};
        \node[file, below=0.1cm of ef1, minimum width=1.5cm] (ef2) {*\_dev.tsv};
        \node[file, below=0.1cm of ef2, minimum width=1.5cm] (ef3) {*\_test.tsv};

        \begin{scope}[on background layer]
            \node[rectangle, rounded corners=10pt, draw=procborder, fill=procbg, line width=1.5pt,
                  minimum width=3.8cm, minimum height=3.5cm] (procbox) at (6.5, 4.5) {};
        \end{scope}

        \node[font=\scriptsize, align=left] at (6.5, 5.2) {1. Load tweet texts};
        \node[font=\scriptsize, align=left] at (6.5, 4.6) {2. Match with labels};
        \node[font=\scriptsize, align=left] at (6.5, 4.0) {3. Normalize disaster type};
        \node[font=\tiny\itshape, text=procborder, align=center] at (6.5, 3.2) {hurricane/fire/flood/earthquake};

        \begin{scope}[on background layer]
            \node[rectangle, rounded corners=8pt, draw=outborder, fill=outbg, line width=1.5pt,
                  minimum width=4.5cm, minimum height=7cm] (outbox) at (11.2, 4.5) {};
        \end{scope}
        \node[font=\normalsize\bfseries, text=outtitle] at (11.2, 7.7) {Processed Dataset};

        \node[folder] (jo) at (11.2, 6.8) {jsonl\_output/};
        \node[file, below=0.15cm of jo] (jo1) {train.jsonl};
        \node[font=\tiny, text=gray, right=0.05cm of jo1] {(53,531)};
        \node[file, below=0.12cm of jo1] (jo2) {dev.jsonl};
        \node[font=\tiny, text=gray, right=0.05cm of jo2] {(7,793)};
        \node[file, below=0.12cm of jo2] (jo3) {test.jsonl};
        \node[font=\tiny, text=gray, right=0.05cm of jo3] {(15,160)};

        \node[rectangle, rounded corners=3pt, draw=outborder, fill=white,
              minimum width=4cm, minimum height=1.2cm, line width=0.8pt] (fieldbox) at (11.2, 3.8) {};
        \node[font=\scriptsize\bfseries, text=outtitle] at (11.2, 4.2) {Output Fields:};
        \node[font=\tiny\ttfamily] at (11.2, 3.7) {tweet\_id | tweet | label};
        \node[font=\tiny\ttfamily] at (11.2, 3.35) {event\_name | event\_type};

        \node[rectangle, rounded corners=4pt, draw=outborder, fill=totalbox, line width=1pt,
              minimum width=3.5cm, minimum height=0.6cm] (totbox) at (11.2, 1.8) {};
        \node[font=\tiny\bfseries, text=outtitle] at (11.2, 1.8) {Total: 76,484 samples};

        \draw[bigarrow] (4.1, 5.5) -- (4.6, 5);
        \draw[bigarrow] (4.1, 3.5) -- (4.6, 4);
        \draw[bigarrow] (8.4, 4.5) -- (8.9, 4.5);

        \end{tikzpicture}
    }
    \caption{HumAID dataset preprocessing pipeline transforming raw TSV files into unified JSONL format with 76,484 samples.}
    \label{fig:preprocessing}
\end{figure}

As illustrated in Figure~\ref{fig:preprocessing}, the preprocessing pipeline transforms raw data---comprising label files in \texttt{all\_combined/} and event-specific texts in \texttt{events\_set1/2/}---into a unified format through three stages: (1) extracting tweet content, (2) aligning texts with humanitarian labels via tweet identifiers, and (3) normalizing event names into four disaster categories (\textit{hurricane, fire, flood, earthquake}). The final processed dataset, stored in \texttt{jsonl\_output/}, consists of 76,484 samples partitioned into \texttt{train} (53,531), \texttt{dev} (7,793), and \texttt{test} (15,160) splits. Each record includes five fields: \texttt{tweet\_id}, \texttt{tweet}, \texttt{label}, \texttt{event\_name}, and \texttt{event\_type}.

\begin{figure}[!t]
    \centering
    \includegraphics[width=\linewidth]{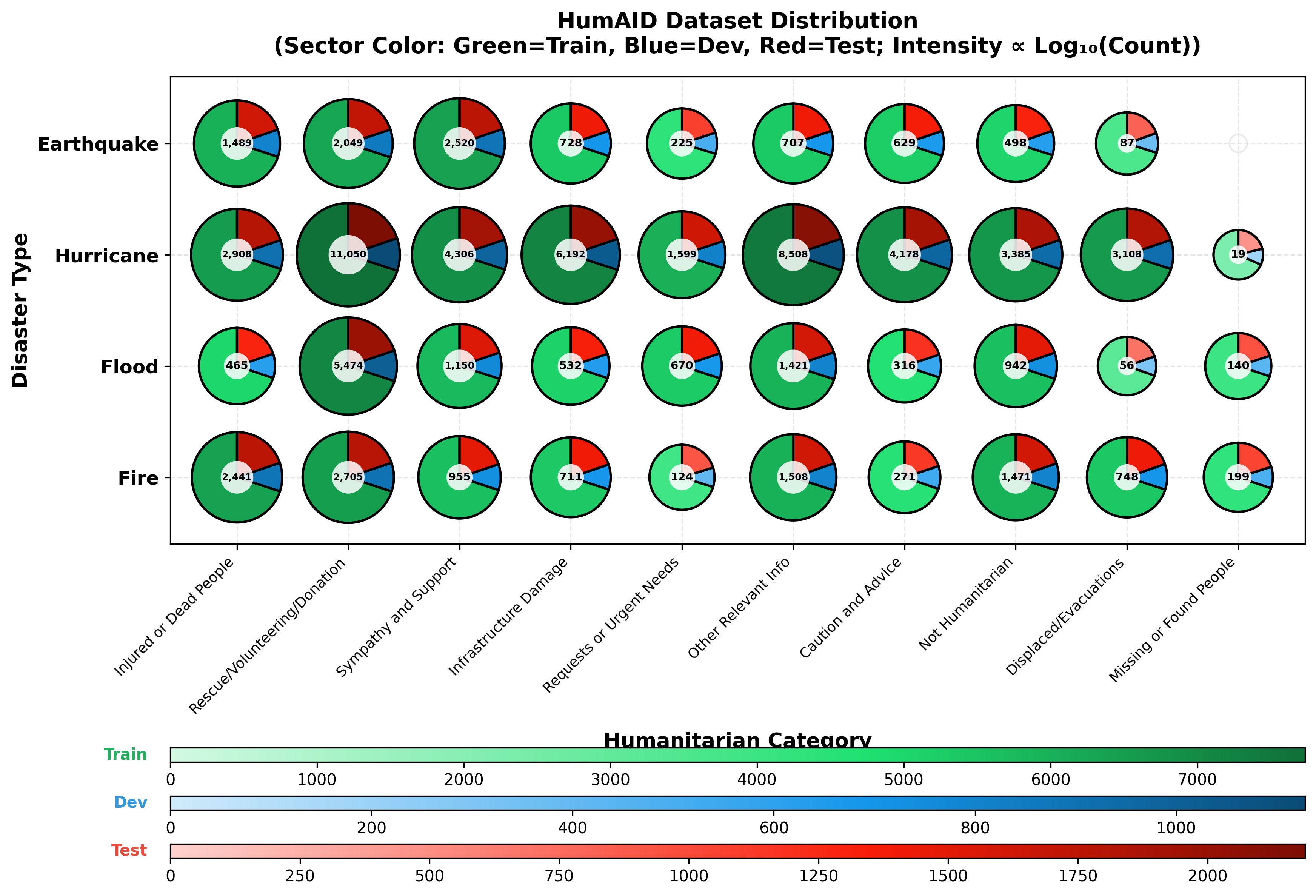}
    \caption{HumAID dataset distribution across humanitarian categories and disaster types. Pie sectors represent train/dev/test splits (green/blue/red); color intensity indicates log-scaled counts; pie size reflects total samples.}
    \label{fig:distribution}
\end{figure}

The following code block illustrates the structure of a single record in the processed JSONL format:

\begin{verbatim}
{
  "tweet_id": "721872405916856321",
  "tweet": "Powerful Ecuador quake kills 
  at least 235...",
  "label": "injured_or_dead_people",
  "event_name": "ecuador_earthquake_2016",
  "event_type": "earthquake"
}
\end{verbatim}

Figure~\ref{fig:distribution} presents a comprehensive visualization of the dataset distribution across three dimensions: humanitarian categories (x-axis), disaster types (y-axis), and data splits (pie chart sectors). This pie-heatmap matrix reveals several important characteristics of the HumAID dataset that significantly influence downstream classification tasks.


\paragraph{Category Imbalance}
The dataset is highly imbalanced: \textit{Rescue/Volunteering/Donation} dominates (21{,}278; 27.8\%), followed by \textit{Other Relevant Information} (12{,}144; 15.9\%) and \textit{Sympathy and Support} (8{,}931; 11.7\%), whereas \textit{Missing or Found People} is rare (358; 0.5\%), yielding a $\sim$59:1 most--least ratio. Such imbalance can bias classifiers toward majority classes and reduce minority-class recall.

\paragraph{Cross-Disaster Variability}
Category distributions differ markedly by disaster type: hurricane events contribute the largest share across most categories, especially \textit{Rescue/Volunteering/Donation} (11{,}050) and \textit{Other Relevant Information} (8{,}508); earthquakes concentrate relatively more in \textit{Sympathy and Support} (2{,}520); and fires are elevated in \textit{Injured or Dead People} (2{,}441). This suggests potential distribution shifts when generalizing across scenarios.

\paragraph{Split Consistency}
Train/dev/test splits remain proportionally consistent (approximately 70\%/10\%/20\%) across most category--disaster combinations, supporting reliable evaluation of generalization.

\paragraph{Implications for Research}
These properties motivate: (1) emphasizing macro-averaged metrics (e.g., macro-F1) over accuracy; (2) considering augmentation or transfer learning for sparse categories such as \textit{Missing or Found People}; and (3) recognizing that semantic overlap between \textit{Not Humanitarian} and \textit{Other Relevant Information}, alongside their large sample sizes, may exacerbate confusion.

\section{Large Language Models for Humanitarian Classification}
\label{sec:llm_background}

This section briefly summarizes the modeling background needed for the progressive experiments in this research. Modern large language models (LLMs) are built upon the Transformer architecture \cite{vaswani2017attention} and are typically trained as decoder-only, autoregressive models. This formulation naturally supports prompt-based inference, which motivates our evaluation pipeline from zero-shot and few-shot prompting to retrieval-augmented prompting and parameter-efficient fine-tuning.

\paragraph{Transformer self-attention.}
The Transformer replaces recurrence with self-attention, enabling each token to aggregate context from the entire sequence \cite{vaswani2017attention}. The core computation is scaled dot-product attention:
\begin{equation}
\mathrm{Attention}(\mathbf{Q}, \mathbf{K}, \mathbf{V})
=
\mathrm{softmax}\!\left(\frac{\mathbf{Q}\mathbf{K}^\top}{\sqrt{d_k}}\right)\mathbf{V},
\label{eq:attention}
\end{equation}
where $\mathbf{Q}$, $\mathbf{K}$, and $\mathbf{V}$ denote the query, key, and value projections. Multi-head attention applies this operation in multiple subspaces and then concatenates the results to form richer contextual representations \cite{vaswani2017attention}.

\paragraph{Autoregressive language modeling and in-context evaluation}
Decoder-only LLMs are commonly trained with the next-token prediction objective:
\begin{equation}
\mathcal{L}
=
-\sum_{t=1}^{T}\log P(x_t \mid x_{<t};\theta),
\label{eq:lm_objective}
\end{equation}
which enables instruction-style prompting and few-shot demonstrations at inference time \cite{brown2020language}. In this research, such prompting-based settings serve as the initial baselines for humanitarian tweet classification before introducing stronger adaptation methods.

\paragraph{Backbone choice}
Building on open-weight LLM families, this research adopts LLaMA~3.1~8B-Instruct \cite{dubey2024llama3} as a practical cost--capability trade-off for controlled label generation and efficient adaptation. The following section details the model-specific components (e.g., normalization, positional encoding, and gating) and then reports results progressively for zero-shot prompting, few-shot prompting, retrieval-augmented prompting, and LoRA-based fine-tuning.

\paragraph{LLaMA 3.1 8B Architecture}
For this study, we employ LLaMA~3.1~8B-Instruct~\cite{dubey2024llama3} as the backbone model, chosen as a practical cost--capability trade-off for humanitarian tweet classification. The model follows a decoder-only Transformer design and incorporates several widely used components:

\begin{itemize}
    \item \textbf{Grouped-Query Attention (GQA)}: improves inference efficiency by sharing key--value heads across multiple query heads, reducing KV-cache cost while retaining multi-head querying behavior.
    
    \item \textbf{RMSNorm}: replaces LayerNorm with root-mean-square normalization \cite{zhang2019rmsnorm}:
    \begin{equation}
    \mathrm{RMSNorm}(\mathbf{x}) =
    \frac{\mathbf{x}}{\sqrt{\frac{1}{d}\sum_{i=1}^{d}x_i^2+\epsilon}}
    \odot \boldsymbol{\gamma},
    \label{eq:rmsnorm}
    \end{equation}
    where $\boldsymbol{\gamma}\in\mathbb{R}^{d}$ is a learnable scaling vector and $\epsilon$ is a small constant for numerical stability.
    
    \item \textbf{RoPE (Rotary Position Embedding)}: injects positional information into attention by rotating query/key vectors as a function of token position, supporting relative-position modeling in decoder-only Transformers \cite{su2024roformer}.
    
    \item \textbf{SwiGLU activation}: uses a gated feed-forward formulation to improve expressivity \cite{shazeer2020glu}:
    \begin{equation}
    \mathrm{SwiGLU}(\mathbf{x}) =
    \mathrm{Swish}(\mathbf{x}\mathbf{W}_1)\odot(\mathbf{x}\mathbf{W}_2).
    \label{eq:swiglu}
    \end{equation}
\end{itemize}

\paragraph{Suitability for Humanitarian Classification}
The selection of LLaMA~3.1~8B-Instruct is motivated by task and deployment considerations:

\begin{enumerate}
    \item \textbf{Instruction interface for constrained outputs}: the instruct-tuned format provides a natural prompting interface for generating a fixed label schema, although raw zero-/few-shot performance on this task is limited and motivates adaptation.
    \item \textbf{Efficiency at the 8B scale}: the model size enables iterative experimentation and cost-aware evaluation under academic compute budgets.
    \item \textbf{Compatibility with parameter-efficient adaptation}: LoRA can specialize the model to humanitarian labels while updating only a small subset of parameters, making it suitable for progressive evaluation from prompting baselines to fine-tuned settings.
\end{enumerate}

Building on this background, the next section presents the experimental design and reports results progressively for zero-shot prompting, few-shot prompting, retrieval-augmented prompting, and LoRA-based fine-tuning.

\section{Method}

All experiments were conducted on a single NVIDIA GeForce RTX 3090 GPU with 24GB VRAM. Our software environment consists of PyTorch 2.9.0 with CUDA 12.8, Hugging Face Transformers 4.46.0, PEFT 0.13.2 for parameter-efficient fine-tuning, and TRL 0.12.0 for supervised fine-tuning. For high-throughput inference, we utilize vLLM 0.12.0 with continuous batching. All models are loaded in BFloat16 precision unless otherwise specified.

\subsection{Baseline Prompting Strategies}
\label{sec:prompting}

Our research addresses a \textbf{dual-task multi-class classification problem}: we simultaneously classify disaster-related tweets into (1) humanitarian information categories and (2) disaster event types. This joint formulation supports a more complete understanding of crisis communications, enabling both humanitarian response coordination and event-specific analysis.

Before exploring parameter-efficient fine-tuning, we evaluate the base Llama~3.1~8B-Instruct model under a sequence of prompting strategies. We emphasize \textbf{prompt constraints} because unconstrained generation frequently violates the required label schema (e.g., free-form explanations or non-standard labels), making outputs unusable for automated evaluation and downstream pipelines.

\subsubsection{Task Formulation}
\label{subsec:task_formulation}

\begin{table}[!t]
\centering
\caption{Dual-task classification targets in our system.}
\label{tab:dual_task}
\footnotesize 
\setlength{\tabcolsep}{2pt}

\newcommand{\slashunderscore}{\allowbreak\_}

\begin{tabularx}{\columnwidth}{lc >{\raggedright\arraybackslash}X}
\toprule
\textbf{Task} & \textbf{\#Cls} & \textbf{Categories} \\
\midrule
\textbf{Hum.} & 10 & 
\texttt{caution\_and\_advice, 
displaced\_people\_and\_evacuations, 
infrastructure\_and\_utility\_damage, 
injured\_or\_dead\_people, 
missing\_or\_found\_people, 
not\_humanitarian, 
other\_relevant\_information, 
requests\_or\_urgent\_needs, 
rescue\_volunteering\_or\_donation\_effort, 
sympathy\_and\_support} \\
\midrule
\textbf{Event} & 4 & \texttt{earthquake, fire, flood, hurricane} \\
\bottomrule
\end{tabularx}
\end{table}

Given a tweet $x$, our objective is to predict two categorical labels:
\begin{equation}
    f(x) \rightarrow (y_h, y_e),
\end{equation}
where $y_h \in \mathcal{Y}_h$ is the humanitarian category and $y_e \in \mathcal{Y}_e$ is the event type. Table~\ref{tab:dual_task} summarizes the label spaces used in this research.

\subsubsection{Prompt Constraints and Baseline Setups}
\label{subsec:prompt_constraints}

LLMs are sensitive to prompt format. Without explicit constraints, we frequently observe \emph{schema-violating generations}, including (i) output format violations (free-form text), (ii) input repetition, and (iii) label-set violations (non-standard labels). To mitigate these issues, all baselines in this section use a \textbf{well-constrained prompt} that (1) enumerates the full label set, (2) requires a \textbf{single JSON object} with fixed field names, and (3) prohibits explanations. \textbf{The exact zero-shot prompt template used in the main experiments is provided in Box~\ref{box:zero_shot_prompt}.}

\begin{tcolorbox}[colback=gray!5, colframe=gray!50, title=Zero-Shot Prompt Template (Main Experiments), label=box:zero_shot_prompt]
\small
\texttt{You are an expert disaster tweet classifier.}\\
\texttt{You must classify each tweet into TWO fields:}\\[0.4em]
\texttt{1) Humanitarian Label (choose exactly ONE):}\\
\texttt{caution\_and\_advice,}\\
\texttt{displaced\_people\_and\_evacuations,}\\
\texttt{infrastructure\_and\_utility\_damage,}\\
\texttt{injured\_or\_dead\_people,}\\
\texttt{missing\_or\_found\_people,}\\
\texttt{not\_humanitarian,}\\
\texttt{other\_relevant\_information,}\\
\texttt{requests\_or\_urgent\_needs,}\\
\texttt{rescue\_volunteering\_or\_donation\_effort,}\\
\texttt{sympathy\_and\_support}\\[0.4em]
\texttt{2) Event Type (choose exactly ONE):}\\
\texttt{earthquake, fire, flood, hurricane}\\[0.6em]
\texttt{Return ONLY ONE JSON object. No explanation.}\\
\texttt{Use this EXACT format:}\\
\texttt{\{"humanitarian\_label": "...", "event\_type": "..."\}}\\[0.6em]
\texttt{Tweet: \{tweet\}}
\end{tcolorbox}

We evaluate the following baselines:
\begin{itemize}
    \item \textbf{Zero-shot prompting}: direct classification without demonstrations.
    \item \textbf{Few-shot prompting}: $k \in \{5,10\}$ labeled demonstrations are prepended to the test instance.
We consider multiple demonstration construction strategies, including (i) \textit{static stratified sampling} to encourage label coverage, (ii) \textit{naive retrieval-augmented demonstrations} based on TF-IDF cosine similarity, and (iii) \textit{LLM-generated demonstrations} created under the same label schema and constraints.
Implementation details and templates are provided in Appendix~\ref{app:prompts} (Appendix~\ref{app:fewshot_examples}).
\end{itemize}

\paragraph{Inference configuration}
We use deterministic decoding (greedy; $\text{temperature}=0$, $\text{top\_p}=1.0$) with a small generation budget (\texttt{max\_tokens}=50), which is sufficient for the required JSON output.

\subsubsection{Output Validation and Evaluation}
\label{subsec:eval_metrics}

To ensure a fair comparison across baselines, we apply \textbf{format-only} validation: we directly parse the model output as a single JSON object. We report standard classification metrics (macro-/weighted-F1 and accuracy) for valid outputs.

\begin{align}
    \text{Precision} &= \frac{TP}{TP + FP} \\
    \text{Recall} &= \frac{TP}{TP + FN} \\
    \text{F1} &= \frac{2 \cdot \text{Precision} \cdot \text{Recall}}{\text{Precision} + \text{Recall}}
\end{align}

Accuracy is computed separately for the humanitarian task and the event-type task:
\begin{equation}
\text{Accuracy}_h = \frac{|\{i : \hat{y}_{h,i} = y_{h,i}\}|}{N}, \quad
\text{Accuracy}_e = \frac{|\{i : \hat{y}_{e,i} = y_{e,i}\}|}{N}.
\end{equation}

\subsection{Parameter-Efficient Fine-Tuning with LoRA}
\label{subsec:lora_finetuning}

While prompting strategies provide a zero-shot baseline for disaster tweet classification, achieving higher accuracy requires task-specific adaptation of the language model. Traditional full-parameter fine-tuning of Llama 3.1 8B (approximately 8 billion parameters) presents significant computational challenges, requiring substantial GPU memory and extended training time. To address these resource constraints, we adopt \textbf{Low-Rank Adaptation (LoRA)} as our primary fine-tuning approach.

A key design principle of our approach is \textbf{maintaining prompt consistency} between zero-shot inference and fine-tuning. We employ Supervised Fine-Tuning (SFT) using the \textbf{same prompt template} as described before. 
This prompt consistency ensures:
\begin{enumerate}
    \item \textbf{Seamless Transition}: Models can switch between zero-shot and fine-tuned inference without prompt modification
    \item \textbf{Fair Comparison}: Performance differences originate from fine-tuning rather than prompt variations
    \item \textbf{Reproducibility}: Identical prompts enable direct comparison across different training strategies
\end{enumerate}

\subsubsection{Motivation for LoRA Adoption}

Recent research demonstrates that LoRA provides an excellent trade-off between performance and computational efficiency. According to \cite{yinCrisisSenseLLMInstructionFineTuned2024}, LoRA-adapted models can achieve up to \textbf{96.7\% of full-parameter fine-tuning performance} on disaster humanitarian tweet classification tasks while training only a fraction of parameters. This finding is particularly significant for crisis informatics applications where:

\begin{itemize}
    \item \textbf{Resource Constraints}: Emergency response organizations often lack access to high-end computing infrastructure
    \item \textbf{Rapid Deployment}: Disaster scenarios require quick model adaptation to new event types
    \item \textbf{Cost Efficiency}: Reduced training costs enable broader adoption of AI-assisted crisis response
\end{itemize}

This cost-performance balance aligns with our research objective: achieving high classification accuracy with minimal computational resources.

The parameter efficiency of LoRA is substantial. For Llama 3.1 8B with rank-32 configuration:

\begin{equation}
    \text{Trainable Ratio} = \frac{\text{LoRA Parameters}}{\text{Total Parameters}} = \frac{167M}{8B} \approx 2.09\%
\end{equation}

This means we train only approximately 2\% of the model parameters while achieving performance that approaches full fine-tuning.

\subsubsection{LoRA Technical Background}

LoRA freezes the pre-trained model weights and injects trainable low-rank decomposition matrices into each layer of the Transformer architecture. For a pre-trained weight matrix $W_0 \in \mathbb{R}^{d \times k}$, LoRA represents the weight update as:

\begin{equation}
    W = W_0 + \Delta W = W_0 + BA
\end{equation}

where $B \in \mathbb{R}^{d \times r}$ and $A \in \mathbb{R}^{r \times k}$, with the rank $r \ll \min(d, k)$. During training, $W_0$ remains frozen while $A$ and $B$ are optimized. The scaling factor is applied as:

\begin{equation}
    h = W_0 x + \frac{\alpha}{r} BAx
\end{equation}

where $\alpha$ is the LoRA scaling hyperparameter (lora\_alpha).

\subsubsection{Implementation Configuration}

We implement LoRA fine-tuning using the PEFT (Parameter-Efficient Fine-Tuning) library with the TRL (Transformer Reinforcement Learning) framework. Table~\ref{tab:lora_config} summarizes our LoRA configurations.

\begin{table}[!t]
\begin{center}
\caption{LoRA fine-tuning configurations for Llama 3.1 8B}
\label{tab:lora_config}
\begin{tabular}{lcc}
\toprule
\textbf{Parameter} & \textbf{Standard LoRA} & \textbf{High-Rank LoRA} \\
\midrule
Rank ($r$) & 32 & 64 \\
Alpha ($\alpha$) & 64 & 128 \\
Dropout & 0.05 & 0.05 \\
Bias & none & none \\
\midrule
\multicolumn{3}{c}{\textbf{Target Modules}} \\
\midrule
\multicolumn{3}{p{8cm}}{\texttt{q\_proj, k\_proj, v\_proj, o\_proj, gate\_proj, up\_proj, down\_proj}}\\
\bottomrule
\end{tabular}
\end{center}
\end{table}

Our configuration targets all attention projection layers (query, key, value, output) and all feedforward network layers (gate, up, down projections). This comprehensive coverage ensures that the adapter can learn task-specific representations while maintaining the general language understanding capabilities of the base model.

\paragraph{Training Hyperparameters:}
\begin{itemize}
    \item Learning rate: $2 \times 10^{-4}$ with cosine scheduler
    \item Batch size: $4 \times 4$ (with gradient accumulation)
    \item Training epochs: 3
    \item Max sequence length: 512 tokens
    \item Warmup ratio: 0.03
    \item Optimizer: AdamW with weight decay 0.01
\end{itemize}

To further reduce memory requirements, we also experiment with \textbf{QLoRA} (Quantized LoRA), which combines 4-bit quantization with LoRA adaptation. QLoRA uses the NormalFloat4 (NF4) quantization scheme with double quantization enabled, achieving approximately 50\% memory reduction compared to BF16 precision. This makes it particularly valuable when GPU memory is scarce.

\subsection{Retrieval-Augmented Generation (RAG)}
\label{subsec:rag}

\textbf{Retrieval-Augmented Generation (RAG)} \cite{lewis2020retrieval} combines retrieval-based and generation-based methods by first retrieving relevant examples from an external knowledge base, then conditioning the generation on both the input query and retrieved context. For disaster tweet classification, RAG enables the model to reference similar labeled examples during inference, potentially improving accuracy through in-context learning.

In this research, RAG is integrated with both the baseline Llama 3.1 8B model (zero-shot setting) and the LoRA-adapted model. This dual integration enables systematic comparison:
\begin{itemize}
    \item \textbf{Baseline + RAG}: Does retrieval help an untrained model by providing task-relevant context?
    \item \textbf{LoRA + RAG}: Does retrieval provide additional gains to an already fine-tuned model?
\end{itemize}

Implementation uses vLLM \cite{kwon2023efficient} for high-throughput inference with $k=3$ retrieved examples and a 4096-token context window. The RAG prompt template is detailed in Appendix~\ref{app:rag_prompt}.

Our RAG implementation consists of four components: embedding optimization, vector index construction, inference strategies, and LLM integration.

\subsubsection{Embedding Model Optimization}
\label{subsubsec:embedding_optimization}

Pre-trained embedding models like \textbf{MiniLM} \cite{wang2020minilm} and \textbf{Sentence-BERT} \cite{reimers2019sentence} may not produce optimal representations for domain-specific retrieval. To explore whether task-specific adaptation can improve retrieval quality, we fine-tune the embedding model using \textbf{contrastive learning} \cite{chen2020simple}.

We construct training pairs from labeled data: \textbf{positive pairs} share the same humanitarian label (similarity target = 1.0), while \textbf{negative pairs} have different labels (similarity target = 0.0). The model is trained with Cosine Similarity Loss:
\begin{equation}
    \mathcal{L} = \frac{1}{N} \sum_{i=1}^{N} \left( \cos(\mathbf{u}_i, \mathbf{v}_i) - y_i \right)^2
\end{equation}
where $\mathbf{u}_i, \mathbf{v}_i$ are the embedding vectors for a text pair and $y_i \in \{0, 1\}$ indicates positive/negative pairing. This loss minimizes the squared error between predicted cosine similarity and the target label, encouraging \textbf{intra-class compactness} (high similarity for same-label pairs) and \textbf{inter-class separation} (low similarity for different-label pairs).

We quantify embedding quality using two clustering metrics:
\begin{itemize}
    \item \textbf{Separation Ratio}: $R = \bar{d}_{\text{inter}} / \bar{d}_{\text{intra}}$, where higher values indicate better class separability
    \item \textbf{Silhouette Score}: measures cluster cohesion relative to separation, ranging from $-1$ to $1$
\end{itemize}

\begin{figure}[!t]
    \centering
    \includegraphics[width=\linewidth]{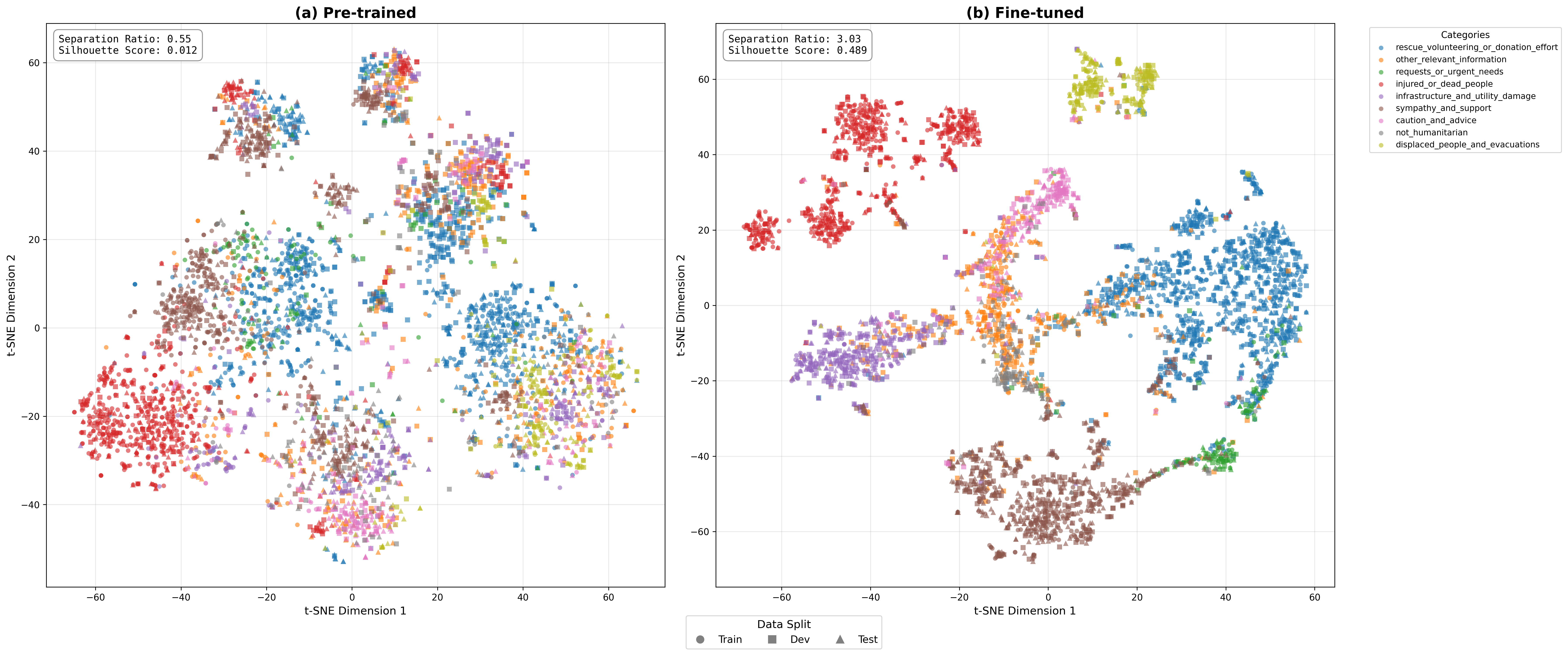}
    \caption{t-SNE visualization of embedding spaces comparing (a) pre-trained and (b) fine-tuned models. Colors indicate humanitarian categories, marker shapes distinguish data splits (circle: train, square: dev, triangle: test). Clustering metrics are shown in the upper-left corner of each subplot.}
    \label{fig:embedding_comparison}
\end{figure}

Figure~\ref{fig:embedding_comparison} visualizes the embedding spaces using t-SNE \cite{van2008visualizing}, comparing the pre-trained and fine-tuned models. Different marker shapes distinguish data splits (circle: train, square: dev, triangle: test). The fine-tuned embedding shows dramatically improved clustering ($R = 3.03$ vs $0.55$; Silhouette = $0.49$ vs $0.01$), indicating successful task adaptation. Training configuration details are provided in Appendix~\ref{app:embedding_config}.

\subsubsection{Vector Index Construction}
\label{subsubsec:index_construction}

We employ \textbf{FAISS} \cite{johnson2019billion} to build a vector index over the training corpus, enabling efficient nearest neighbor retrieval. The indexing pipeline proceeds as follows:

\begin{enumerate}
    \item \textbf{Encoding}: Each training tweet is encoded using the fine-tuned embedding model to produce a 384-dimensional vector.
    \item \textbf{Normalization}: Vectors are L2-normalized to enable cosine similarity via inner product.
    \item \textbf{Indexing}: We use IndexFlatIP (exact inner product search) for maximum retrieval accuracy.
\end{enumerate}

Beyond indexing raw tweet text, we explore a \textbf{label-enriched} indexing strategy: prepending label information to each tweet (formatted as ``Label: [category]. Tweet: [content]''). This strategy embeds classification signals directly into the vector space, enabling retrieval to consider both semantic similarity and label consistency.

\subsubsection{RAG Inference Strategies}
\label{subsubsec:rag_strategies}

We implement three progressively sophisticated inference strategies:

\paragraph{Standard RAG} Every test sample triggers retrieval of top-$k$ semantically similar examples from the training corpus. Retrieved examples are formatted as few-shot demonstrations in the prompt:
\begin{equation}
    P_{\text{RAG}} = P_{\text{system}} \oplus \bigoplus_{i=1}^{k} (\text{Tweet}_i, \text{Label}_i) \oplus \text{Query}
\end{equation}
where $\oplus$ denotes prompt concatenation. This approach provides consistent retrieval but incurs computational overhead for every sample.

\paragraph{Adaptive RAG} A two-phase approach that applies retrieval selectively:
\begin{enumerate}
    \item \textbf{Phase 1}: Direct inference without retrieval; extract confidence score from model output probabilities
    \item \textbf{Phase 2}: For samples with confidence below threshold $\tau$, trigger RAG retrieval and re-inference
\end{enumerate}
This strategy significantly reduces retrieval calls (typically 40-60\% of samples exceed $\tau$) while maintaining accuracy on uncertain cases.

\paragraph{Hybrid Arbitration} Our most sophisticated strategy that intelligently reranks retrieved examples based on consistency with the initial prediction. Given Phase-1 prediction $\hat{y}^{(1)}$ and retrieved neighbors $\mathcal{N}$:
\begin{itemize}
    \item \textit{Consistency Protection}: If neighbors supporting $\hat{y}^{(1)}$ exist, prioritize them to reinforce confident predictions
    \item \textit{Strong Correction}: If $\geq$50\% of neighbors share a different label, use that dominant label's examples to challenge potentially wrong predictions
    \item \textit{Graceful Fallback}: When neighbor labels are ambiguous, revert to original semantic similarity ranking
\end{itemize}
The complete algorithm is provided in Appendix~\ref{app:hybrid_algorithm}.

\section{Experimental Results}
\label{sec:results}

This section presents comprehensive experimental results comparing baseline prompting strategies, LoRA fine-tuning, and RAG augmentation for disaster tweet classification.

\subsection{Baseline Prompting Results}
\label{subsec:baseline_results}

Table~\ref{tab:baseline_overall} summarizes the performance of four baseline prompting strategies. The results reveal a clear hierarchy: \textbf{Dynamic Few-shot} achieves the best performance, followed by Static, Zero-shot, and surprisingly, Manual Few-shot performs worst.

\begin{table}[!t] 
\centering        
\caption{Overall performance comparison of baseline prompting strategies (4 decimal precision).}
\label{tab:baseline_overall}

\resizebox{\columnwidth}{!}{%
\begin{tabular}{lcccccc}
\toprule
\multirow{2}{*}{\textbf{Method}} & \multicolumn{3}{c}{\textbf{Humanitarian}} & \multicolumn{3}{c}{\textbf{Event Type}} \\

\cmidrule(lr){2-4} \cmidrule(lr){5-7}
& Acc. & M-F1 & W-F1 & Acc. & M-F1 & W-F1 \\ 

\midrule
Zero-shot        & 0.4183 & 0.3432 & 0.4228 & 0.6274 & 0.5986 & 0.7456 \\
Manual Few-shot  & 0.3272 & 0.3492 & 0.3823 & 0.4665 & 0.4906 & 0.6201 \\
Static Few-shot  & 0.5051 & 0.4504 & 0.5217 & 0.8197 & 0.7116 & 0.8750 \\
\textbf{Dynamic Few-shot} & \textbf{0.6410} & \textbf{0.5625} & \textbf{0.6534} & \textbf{0.8569} & \textbf{0.7151} & \textbf{0.8893} \\
\bottomrule
\end{tabular}%
}
\end{table}

\subsubsection{Overall Performance Comparison}

Table~\ref{tab:baseline_overall} presents the comprehensive comparison of four baseline prompting strategies. The results reveal a clear performance hierarchy: \textbf{Dynamic Few-shot} achieves the best performance across all metrics, followed by Static Few-shot, Zero-shot, and surprisingly, Manual Few-shot performs worst.

\begin{figure}[!t]
    \centering
    \includegraphics[width=\linewidth]{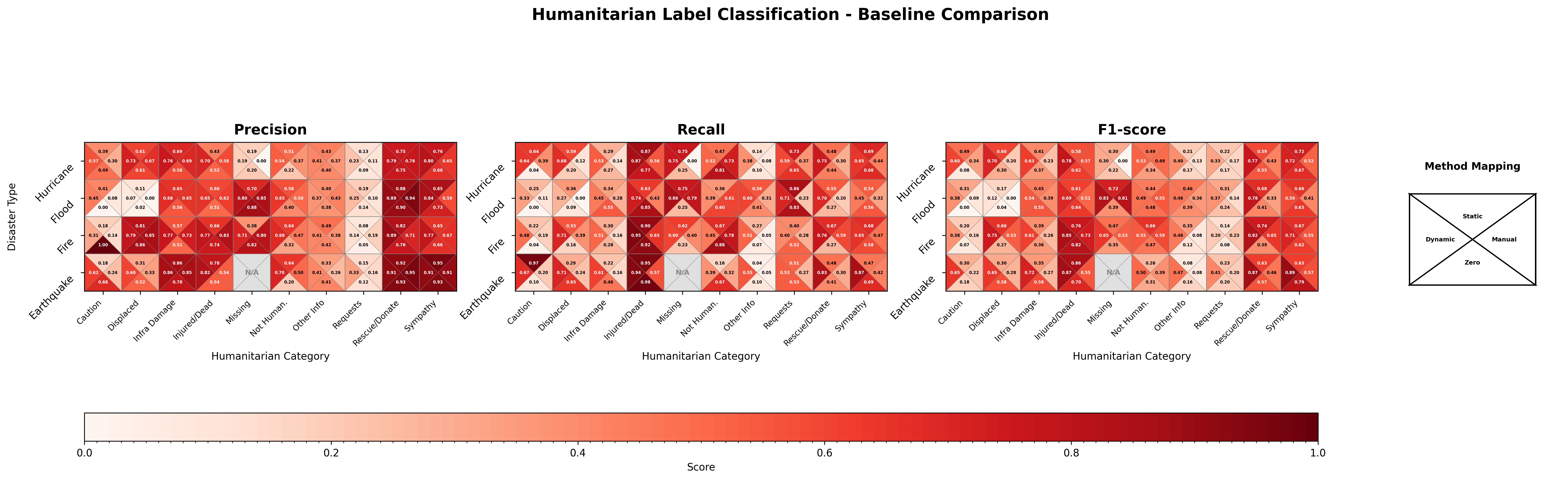}
    \caption{Humanitarian label classification performance by disaster type and category. Each cell shows four triangles (Zero-shot: bottom, Manual: right, Static: top, Dynamic: left). Deeper red indicates higher scores. The N/A cell indicates no test samples exist for that combination (earthquake $\times$ missing\_or\_found\_people).}
    \label{fig:baseline_humanitarian}
\end{figure}

\begin{figure}[!t]
    \centering
    \includegraphics[width=\columnwidth]{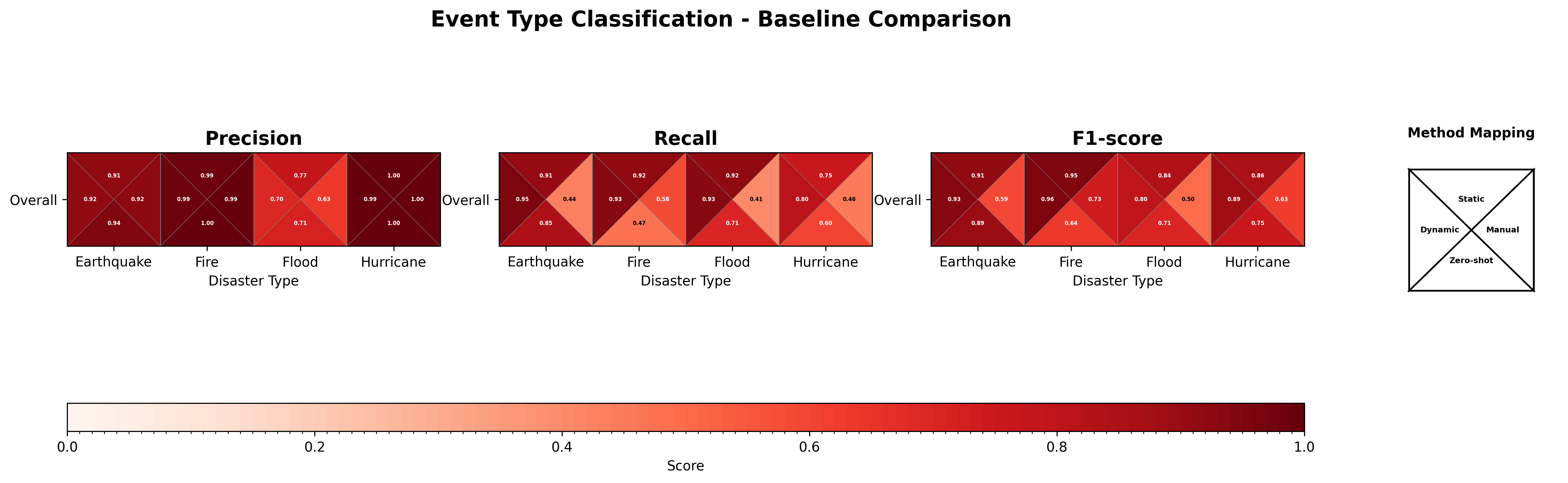}
    \caption{Event type classification performance across four disaster categories. Dynamic Few-shot consistently achieves the highest scores across all event types.}
    \label{fig:baseline_event}
\end{figure}

Figure~\ref{fig:baseline_humanitarian} and Figure~\ref{fig:baseline_event} provide fine-grained visualizations of precision, recall, and F1-score across all categories. Each cell contains four triangles representing the four methods, with color intensity indicating score magnitude.

\paragraph{Humanitarian Classification Analysis}
\begin{itemize}
    \item \textbf{Zero-shot} (Accuracy: 41.83\%) demonstrates the model's inherent understanding of disaster-related content without examples. It achieves high recall on ``not\_humanitarian'' (78.23\%) and ``requests\_or\_urgent\_needs'' (68.33\%), but at the cost of high false positive rates (precision of only 25.14\% and 10.19\% respectively).
    \item \textbf{Manual Few-shot} (Accuracy: 32.72\%) \textit{unexpectedly performs worse} than zero-shot. This counter-intuitive result suggests that manually selected examples may introduce biases or fail to represent the data distribution effectively.
    \item \textbf{Static Few-shot} (Accuracy: 50.51\%) shows significant improvement (+8.68\% over zero-shot) with randomly sampled but fixed examples, indicating that diverse examples outweigh hand-picked ``ideal'' cases.
    \item \textbf{Dynamic Few-shot} (Accuracy: 64.10\%) achieves the best baseline performance by retrieving similar examples dynamically. This +22.27\% improvement over zero-shot demonstrates the value of context-aware retrieval.
\end{itemize}

\paragraph{Event Type Classification Analysis}
Event type classification is inherently easier than humanitarian classification due to fewer classes and clearer semantic boundaries. Key observations:
\begin{itemize}
    \item \textbf{Dynamic Few-shot} achieves 85.69\% accuracy, approaching useful operational thresholds.
    \item \textbf{Flood} is the most challenging category with lower precision (69.67\% for Dynamic), likely due to overlap with hurricane-related flooding content.
    \item \textbf{Fire} and \textbf{hurricane} achieve highest precision ($>$98\%), suggesting distinctive linguistic patterns.
\end{itemize}

\subsubsection{Category-Level Performance}

Table~\ref{tab:baseline_per_class} presents detailed per-class metrics for the Dynamic Few-shot method, which achieves the best overall performance.

\begin{table}[!t]
\centering
\caption{Per-class performance of Dynamic Few-shot on humanitarian classification.}
\label{tab:baseline_per_class}
\footnotesize 
\setlength{\tabcolsep}{3pt} 

\begin{tabularx}{\columnwidth}{>{\raggedright\arraybackslash}X ccc}
\toprule
\textbf{Category} & \textbf{Prec.} & \textbf{Rec.} & \textbf{F1} \\
\midrule
\texttt{caution\_and\_advice} & 0.5508 & 0.6178 & 0.5824 \\
\texttt{displaced\_people\_and\_evacuations} & 0.6988 & 0.6785 & 0.6885 \\
\texttt{infrastructure\_and\_utility\_damage} & 0.7640 & 0.5325 & 0.6276 \\
\texttt{injured\_or\_dead\_people} & 0.7454 & 0.9005 & 0.8156 \\
\texttt{missing\_or\_found\_people} & 0.6000 & 0.7083 & 0.6497 \\
\texttt{not\_humanitarian} & 0.5867 & 0.4731 & 0.5238 \\
\texttt{other\_relevant\_information} & 0.4052 & 0.4358 & 0.4199 \\
\texttt{requests\_or\_urgent\_needs} & 0.2345 & 0.6084 & 0.3385 \\
\texttt{rescue\_volunteering\_or\_donation\_effort} & 0.8360 & 0.7433 & 0.7870 \\
\texttt{sympathy\_and\_support} & 0.8396 & 0.6851 & 0.7545 \\
\bottomrule
\end{tabularx}
\end{table}

\paragraph{Challenging Categories.}
Three categories exhibit notably lower performance:
\begin{enumerate}
    \item \textbf{requests\_or\_urgent\_needs} (F1: 0.3385): Despite high recall (60.84\%), precision is only 23.45 \%. The model over-predicts this category, confusing it with ``rescue\_volunteering\_or\_donation\_effort'' which shares semantic overlap.
    \item \textbf{other\_relevant\_information} (F1: 0.4199): This catch-all category contains semantically diverse content that is difficult to characterize with few examples.
    \item \textbf{not\_humanitarian} (F1: 0.5238): Distinguishing non-humanitarian content from subtle humanitarian references requires fine-grained semantic understanding.
\end{enumerate}

The baseline prompting experiments yield several important insights:
\begin{enumerate}
    \item \textbf{Dynamic retrieval outperforms static examples}: Context-aware example selection provides a 22.27\% accuracy improvement over zero-shot, validating retrieval-based approaches.
    \item \textbf{Manual example curation can backfire}: Hand-picked ``ideal'' examples may not generalize well, with Manual Few-shot underperforming even zero-shot by 9.11\%.
    \item \textbf{Event type classification is substantially easier}: With only 4 classes, this task achieves 85.69\% accuracy compared to 64.10\% for 10-class humanitarian classification.
    \item \textbf{Performance ceiling exists for prompting}: Even the best baseline method (Dynamic Few-shot at 64.10\%) falls short of reliability requirements for practical applications, demonstrating that LoRA fine-tuning is essential to achieve usable performance levels.
\end{enumerate}

\subsection{LoRA Fine-tuning Results}
\label{subsec:lora_results}

LoRA fine-tuning dramatically improves classification performance, achieving \textbf{79.62\% accuracy}—a \textbf{37.79\% absolute improvement} over the zero-shot baseline. This section presents comprehensive results for three LoRA configurations: standard LoRA with rank 32 and 64, and QLoRA with 4-bit quantization.

\subsubsection{Training Dynamics and Configuration Comparison}

\begin{figure}[!t]
    \centering
    \includegraphics[width=\linewidth]{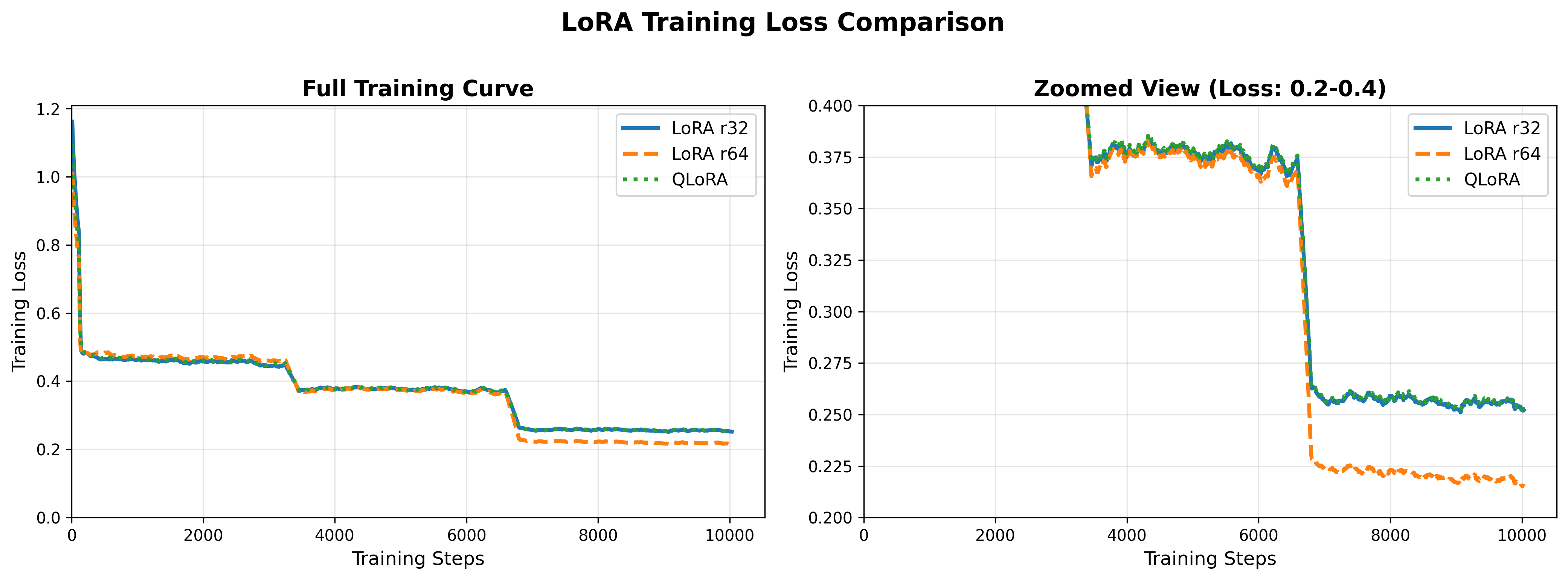}
    \caption{LoRA training loss comparison. Left: Full training curve showing rapid convergence in the first 2000 steps. Right: Zoomed view (loss 0.2--0.4) highlighting differences between configurations.}
    \label{fig:lora_training}
\end{figure}

Figure~\ref{fig:lora_training} shows the training loss curves for all three configurations over approximately 10,000 steps (3 epochs). All configurations converge smoothly with cosine learning rate decay from $2 \times 10^{-4}$ to near zero. The right panel zooms into the 0.2--0.4 loss range, revealing subtle differences: LoRA rank 64 achieves slightly lower final loss, while QLoRA exhibits marginally higher variance due to quantization noise.

Table~\ref{tab:lora_overall} presents the overall performance comparison of LoRA configurations. The results demonstrate that all three achieve comparable accuracy (79--79.6\%), with LoRA rank 64 marginally outperforming rank 32 (+0.11\% accuracy). QLoRA achieves 99.4\% of standard LoRA performance while requiring only half of the memory footprint.

\begin{table}[!t] 
\centering        
\caption{Overall performance comparison of LoRA configurations (4 decimal precision).}
\label{tab:lora_overall}

\resizebox{\columnwidth}{!}{%
\begin{tabular}{lcccccc}
\toprule
\multirow{2}{*}{\textbf{Configuration}} & \multicolumn{3}{c}{\textbf{Humanitarian}} & \multicolumn{3}{c}{\textbf{Event Type}} \\

\cmidrule(lr){2-4} \cmidrule(lr){5-7}
& Acc. & M-F1 & W-F1 & Acc. & M-F1 & W-F1 \\ 

\midrule
LoRA (rank 32) & 0.7951 & 0.7807 & 0.7916 & 0.9875 & 0.9853 & 0.9875 \\
\textbf{LoRA (rank 64)} & \textbf{0.7962} & \textbf{0.7780} & \textbf{0.7932} & \textbf{0.9879} & \textbf{0.9857} & \textbf{0.9879} \\
QLoRA (rank 32) & 0.7942 & 0.7779 & 0.7907 & 0.9875 & 0.9852 & 0.9875 \\
\bottomrule
\end{tabular}%
}
\end{table}

\begin{figure}[!t]
    \centering
    \includegraphics[width=\linewidth]{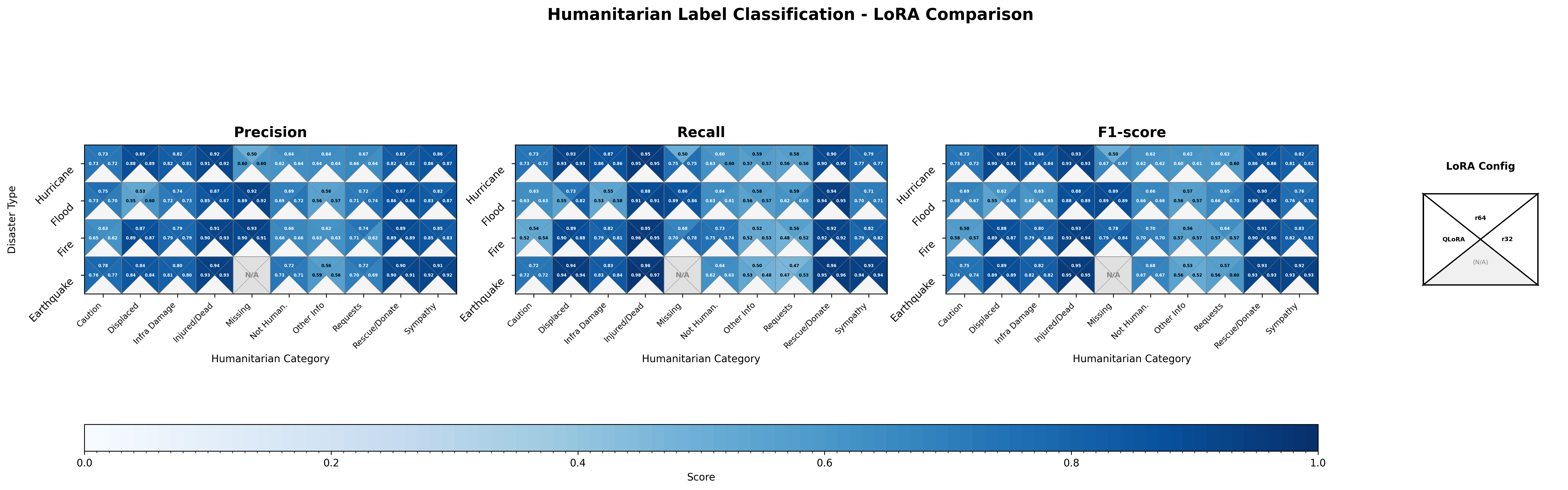}
    \caption{Humanitarian label classification performance by LoRA configuration. Each cell shows three triangles (LoRA r32: right, LoRA r64: top, QLoRA: left). All configurations achieve similar performance, with subtle differences in challenging categories.}
    \label{fig:lora_humanitarian}
\end{figure}

\begin{figure}[!t]
    \centering
    \includegraphics[width=\columnwidth]{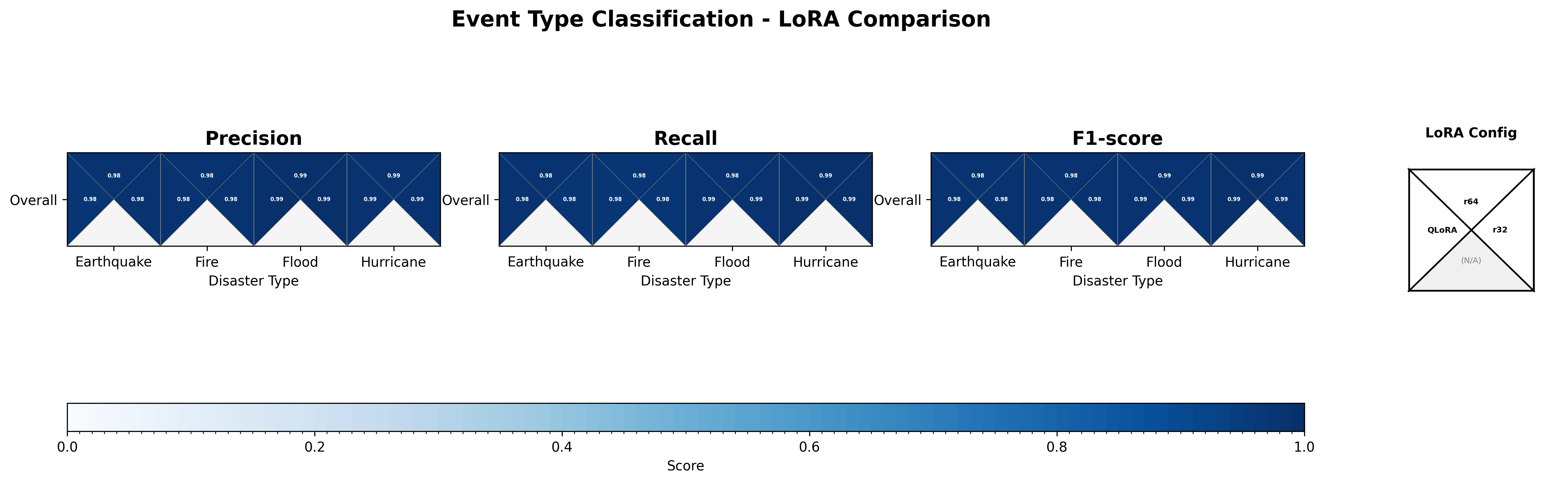}
    \caption{Event type classification achieves near-perfect performance (>98\%) across all LoRA configurations and disaster types.}
    \label{fig:lora_event}
\end{figure}

Figure~\ref{fig:lora_humanitarian} and Figure~\ref{fig:lora_event} visualize the per-category performance using triangle heatmaps. Each cell contains three triangles representing the three LoRA configurations, with deeper blue indicating higher scores.

\subsubsection{Per-Category Performance Analysis}

Table~\ref{tab:lora_per_class} presents the per-class F1 scores for all three configurations. The results reveal consistent patterns: ``injured\_or\_dead\_people'' achieves the highest F1 (0.93+) due to distinctive vocabulary, while ``other\_relevant\_information'' remains the most challenging category (F1 $\approx$ 0.59) due to its semantic diversity.

\begin{table}[!t] 
\centering        
\caption{Per-class F1 scores across LoRA configurations (category names are abbreviated for brevity).}
\label{tab:lora_per_class}
\resizebox{\columnwidth}{!}{%
\begin{tabular}{lccc}
\toprule
\textbf{Category} & \textbf{LoRA r32} & \textbf{LoRA r64} & \textbf{QLoRA} \\
\midrule
Caution \& Advice & 0.7147 & 0.7206 & 0.7230 \\
Displaced \& Evacuations & 0.8960 & 0.8967 & 0.8978 \\
Infrastructure \& Utility & 0.8210 & 0.8240 & 0.8226 \\
Injured or Dead & \textbf{0.9348} & \textbf{0.9325} & \textbf{0.9317} \\
Missing or Found & 0.8467 & 0.8030 & 0.8235 \\
Not Humanitarian & 0.6481 & 0.6507 & 0.6510 \\
Other Relevant Info & 0.5924 & 0.5988 & 0.5913 \\
Requests or Urgent Needs & 0.6222 & 0.6241 & 0.6141 \\
Rescue \& Donation & 0.8839 & 0.8847 & 0.8817 \\
Sympathy \& Support & 0.8469 & 0.8451 & 0.8420 \\
\bottomrule
\end{tabular}%
}
\end{table}

\begin{figure}[!t]
    \centering
    \includegraphics[width=\linewidth]{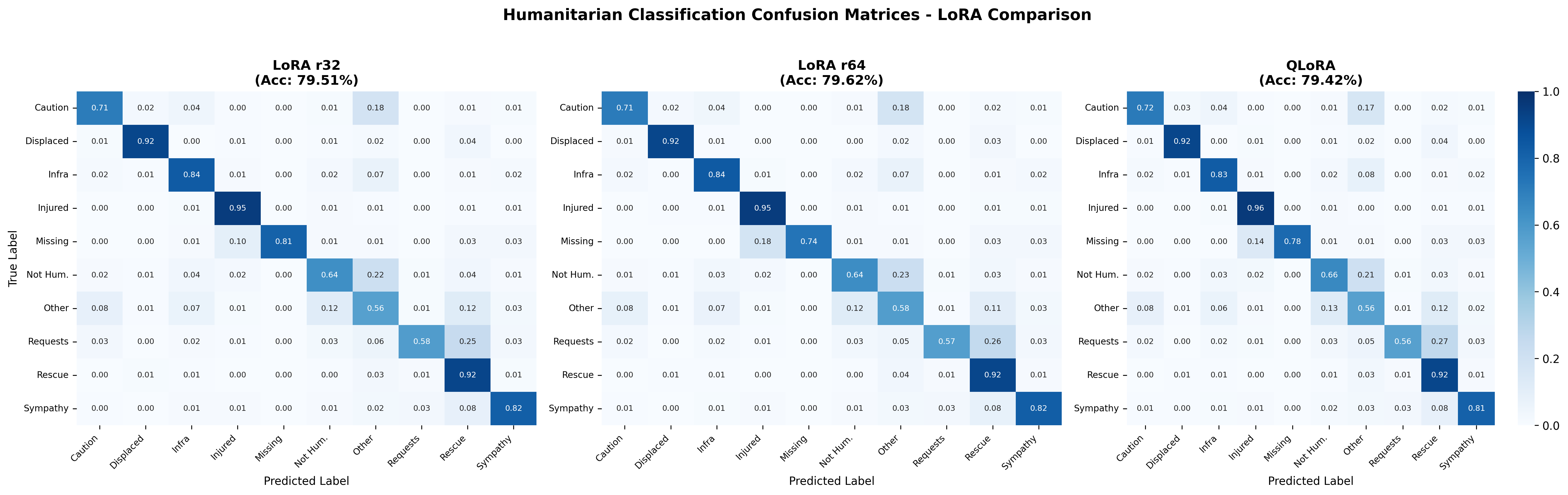}
    \caption{Confusion matrices for humanitarian classification across LoRA configurations. All achieve $\sim$80\% accuracy with similar error patterns.}
    \label{fig:lora_confusion}
\end{figure}

Figure~\ref{fig:lora_confusion} shows the confusion matrices for all three configurations side by side. The diagonal dominance confirms high classification accuracy, while off-diagonal patterns reveal systematic confusions—particularly between ``other\_relevant\_information'' and ``rescue\_volunteering\_or\_donation\_effort''.

\paragraph{Key Findings}
\begin{enumerate}
    \item \textbf{LoRA Superiority}: LoRA improved accuracy from 41.83\% (zero-shot) to 79.51\% (+37.68\%), underscoring the necessity of task-specific fine-tuning for humanitarian classification.
    \item \textbf{Rank Convergence}: Rank has minimal impact; LoRA r64 outperforms r32 by only 0.11\%, suggesting r32 captures sufficient task-specific knowledge. While we report r64 as our best configuration, higher ranks mainly add computational overhead without significant gains.
    \item \textbf{QLoRA Efficiency}: QLoRA retains 99.4\% of standard LoRA performance with 4-bit quantization, providing a viable solution for deployment on resource-constrained hardware.
    \item \textbf{Event Classification Potential}: All configurations achieved $>98.7\%$ accuracy on the 4-class event task. This near-solved state suggests high generalization potential for additional or cascading hazards.
    \item \textbf{Taxonomy Ambiguity}: Performance bottlenecks persist in ``other\_relevant\_info'' (F1=0.59) and ``not\_humanitarian'' (F1=0.65) due to semantic overlap, highlighting the need for targeted data augmentation or refined taxonomies.
\end{enumerate}

\subsection{RAG Augmentation Results}
\label{subsec:rag_results}

This section evaluates RAG's impact on classification performance using the three strategies described in Section~\ref{subsec:rag}: Standard RAG, Adaptive RAG, and Hybrid Arbitration. Since LoRA fine-tuning has essentially solved the event type classification task (accuracy $> 98\%$), this section focuses exclusively on the more challenging \textbf{humanitarian information classification} task.

Table~\ref{tab:rag_overall} presents the comprehensive performance comparison across all RAG configurations, revealing a striking contrast: \textbf{RAG benefits the baseline model but degrades the LoRA model}.

\begin{table}[!t] 
\centering        
\caption{RAG performance comparison across all configurations. Best results per model type are in \textbf{bold}.}
\label{tab:rag_overall}

\resizebox{\columnwidth}{!}{%
\begin{tabular}{llccc}
\toprule
\textbf{Base Model} & \textbf{RAG Configuration} & \textbf{Accuracy} & \textbf{$\Delta$ vs No-RAG} & \textbf{RAG Trigger Rate} \\
\midrule
\multirow{5}{*}{Llama 3.1 8B (ZS)} 
    & No RAG & 0.4183 & — & — \\
    & Pretrained + Std & 0.3823 & −3.60\% & 100\% \\
    & Finetuned + Std & 0.3704 & −4.79\% & 100\% \\
    & Finetuned + Adpt & 0.4918 & +7.35\% & 42.73\% \\
    & Finetuned + Hyb & \textbf{0.5497} & \textbf{+13.14\%} & — \\
\midrule
\multirow{5}{*}{Llama 3.1 8B + LoRA}
    & No RAG & \textbf{0.7951} & — & — \\
    & Pretrained + Std & 0.7742 & −2.09\% & 100\% \\
    & Finetuned + Std & 0.7825 & −1.26\% & 100\% \\
    & Finetuned + Adpt & 0.7901 & −0.50\% & 26.21\% \\
    & Finetuned + Hyb & 0.7919 & −0.32\% & — \\
\bottomrule
\end{tabular}%
}
\end{table}

\begin{figure}[!t]  
    \centering

    \includegraphics[width=\linewidth]{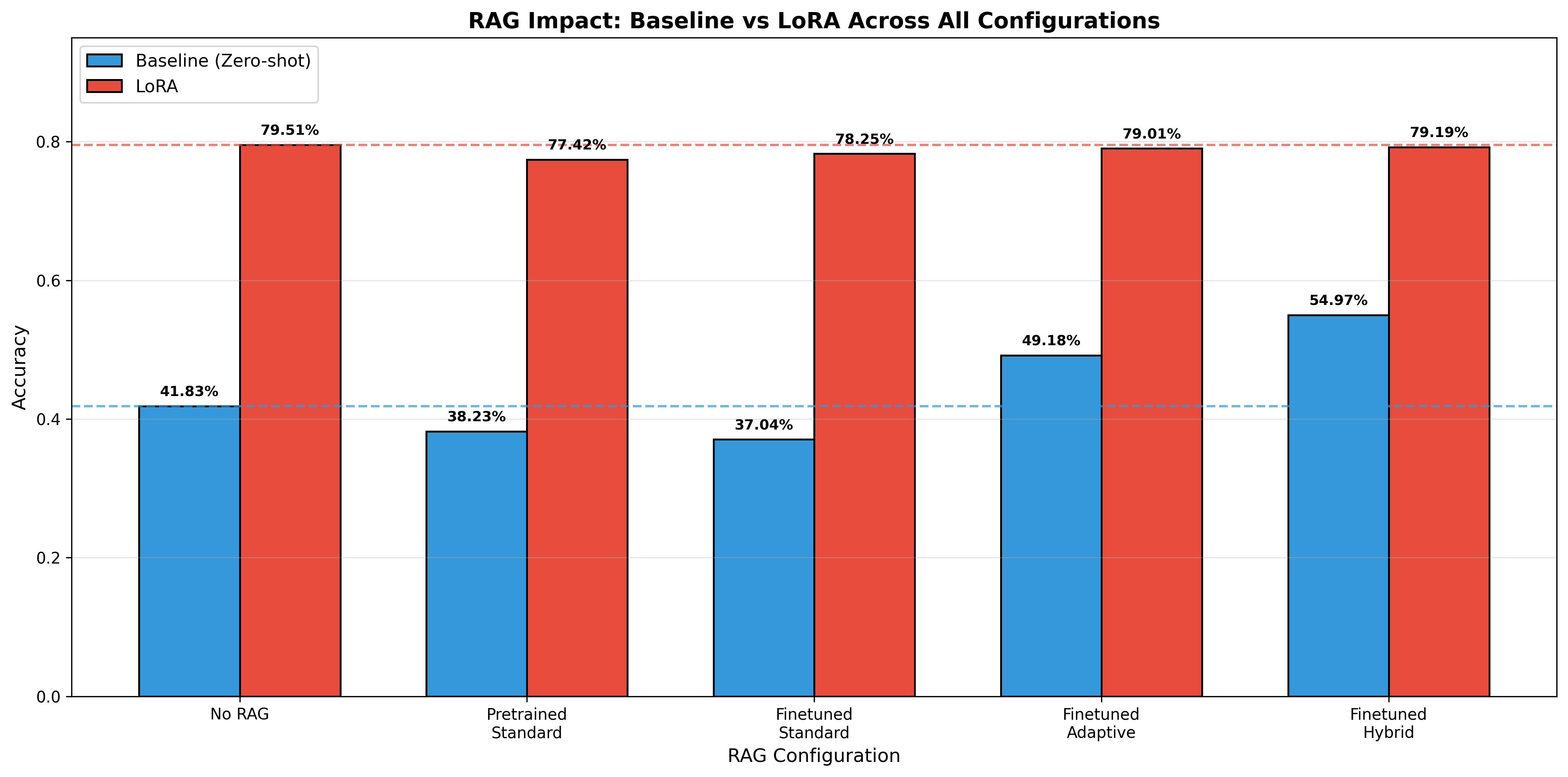}
    \caption{RAG impact comparison: Baseline (zero-shot) vs LoRA across all configurations. Dashed lines indicate the no-RAG baseline for each model type.}
    \label{fig:rag_grouped}
\end{figure}

\begin{figure}[!t]
    \centering
    \includegraphics[width=\linewidth]{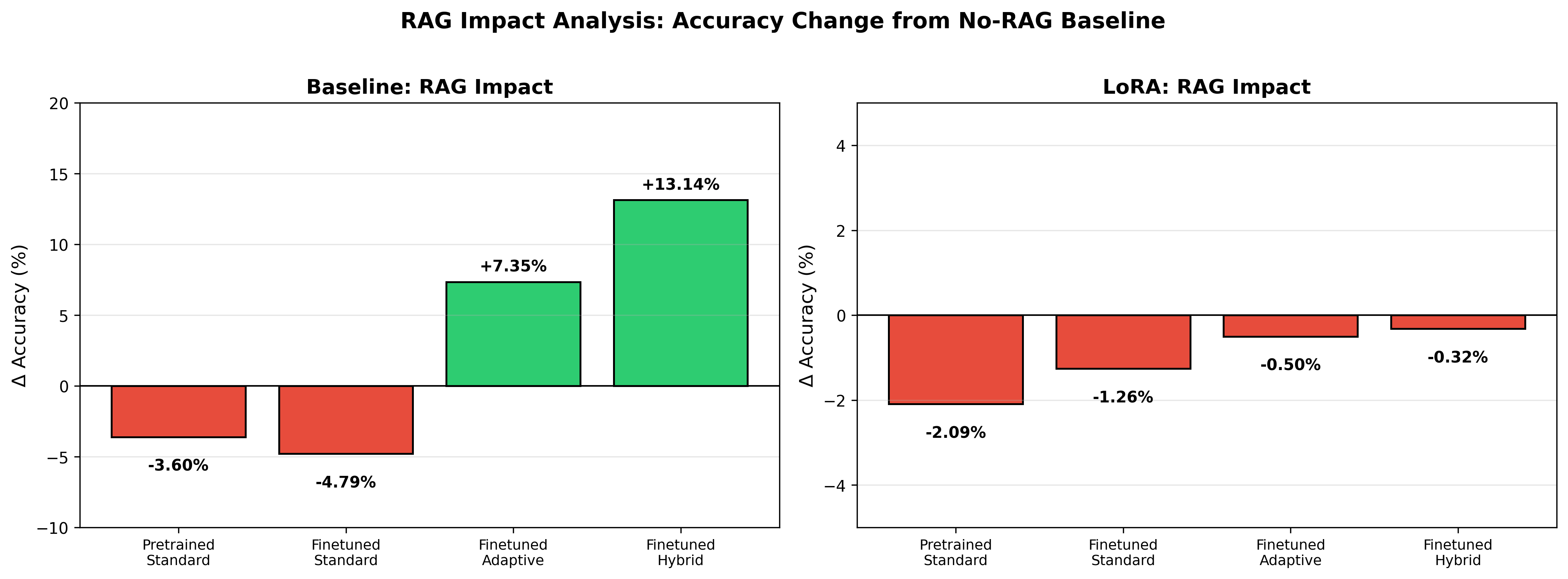}
    \caption{RAG impact analysis: Left panel shows baseline model (green = positive impact), right panel shows LoRA model (red = negative impact).}
    \label{fig:rag_delta}
\end{figure}

For the \textbf{zero-shot baseline}, advanced RAG strategies provide substantial improvements: Standard RAG actually decreases accuracy (−3.6\% to −4.8\%) as the model cannot effectively leverage retrieved examples; Adaptive RAG improves by +7.35\% (trigger rate 42.73\%); Hybrid Arbitration achieves the best result (+13.14\%). However, for the \textbf{LoRA model}, all RAG configurations cause performance degradation: Standard RAG drops −1.3\% to −2.1\%, Adaptive RAG drops −0.50\% (trigger rate only 26.21\%), and even the best Hybrid still drops −0.32\%. Figure~\ref{fig:rag_delta} visualizes this contrasting effect.

\begin{figure}[!t]
    \centering
    \includegraphics[width=\columnwidth]{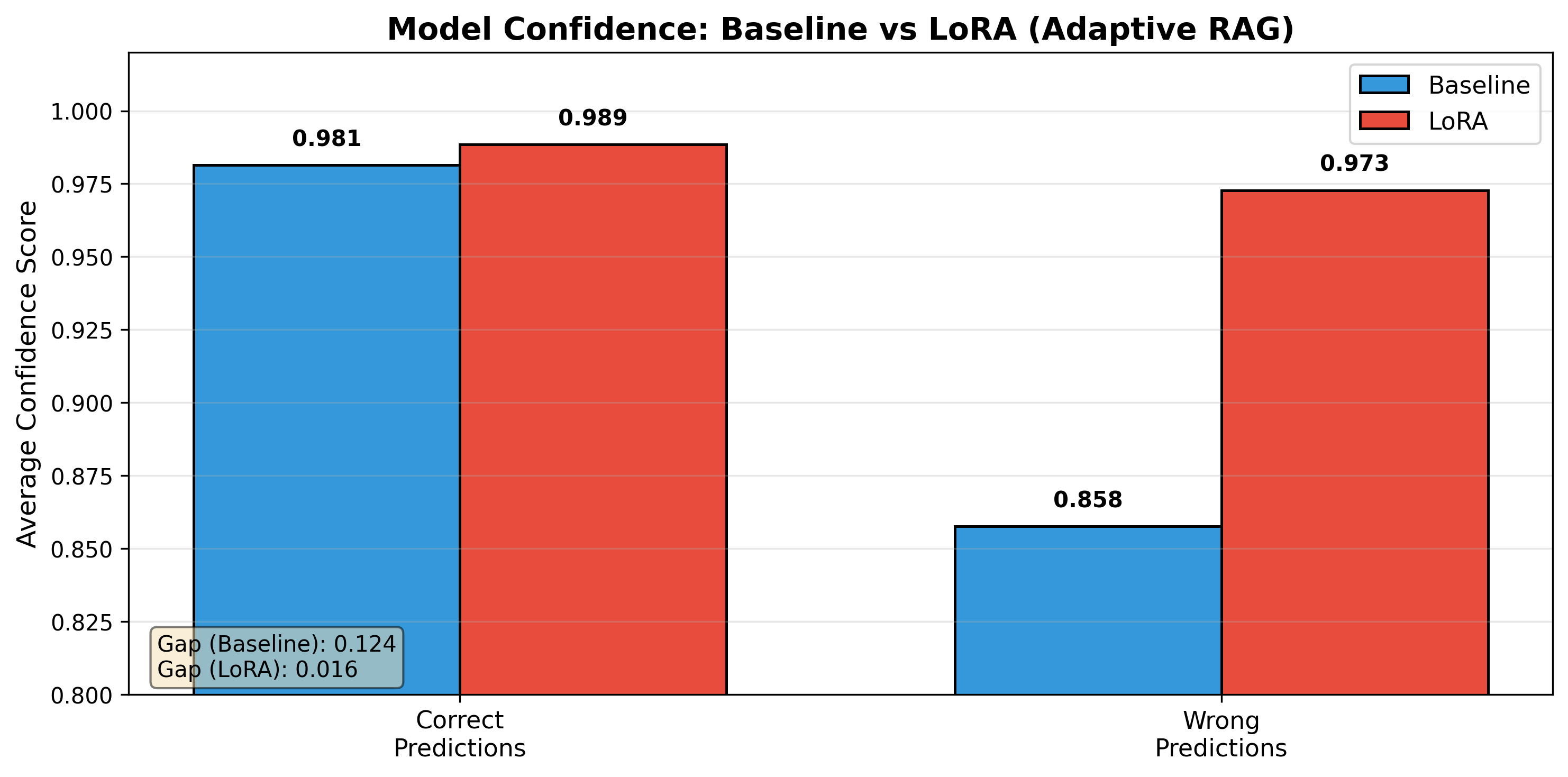}
    \caption{Model confidence comparison under Adaptive RAG. The LoRA model shows higher confidence for both correct and wrong predictions, with a smaller gap (0.016 vs 0.124), making confidence-based filtering less effective.}
    \label{fig:rag_confidence}
\end{figure}

Figure~\ref{fig:rag_confidence}'s confidence analysis reveals the root cause: the baseline model has a clear confidence gap (0.124) between correct (0.981) and wrong (0.858) predictions, enabling effective confidence-based retrieval triggering; the LoRA model has a small confidence gap (0.016), assigning high confidence (0.973) even to wrong predictions—a ``confidently wrong'' phenomenon that renders adaptive filtering ineffective.

To condense this section while preserving the technical logic, we can emphasize the "capability-interference" trade-off: This contrast stems from model capability: while RAG provides essential guidance for zero-shot models, it introduces interference for fine-tuned LoRA models that have already internalized task-specific patterns. In high-performance models, semantic similarity often fails to align with label intent, causing retrieved examples to contradict learned knowledge. \textbf{Core finding}: RAG effectiveness is inversely correlated with model capability—offering a knowledge supplement for weak models but interference noise for strong ones. Thus, RAG is better suited for data-scarce, knowledge-intensive tasks rather than well-defined classification with sufficient training data.

\subsection{Summary}
\label{subsec:results_summary}

\begin{figure}[!t]
    \centering
    \includegraphics[width=\columnwidth]{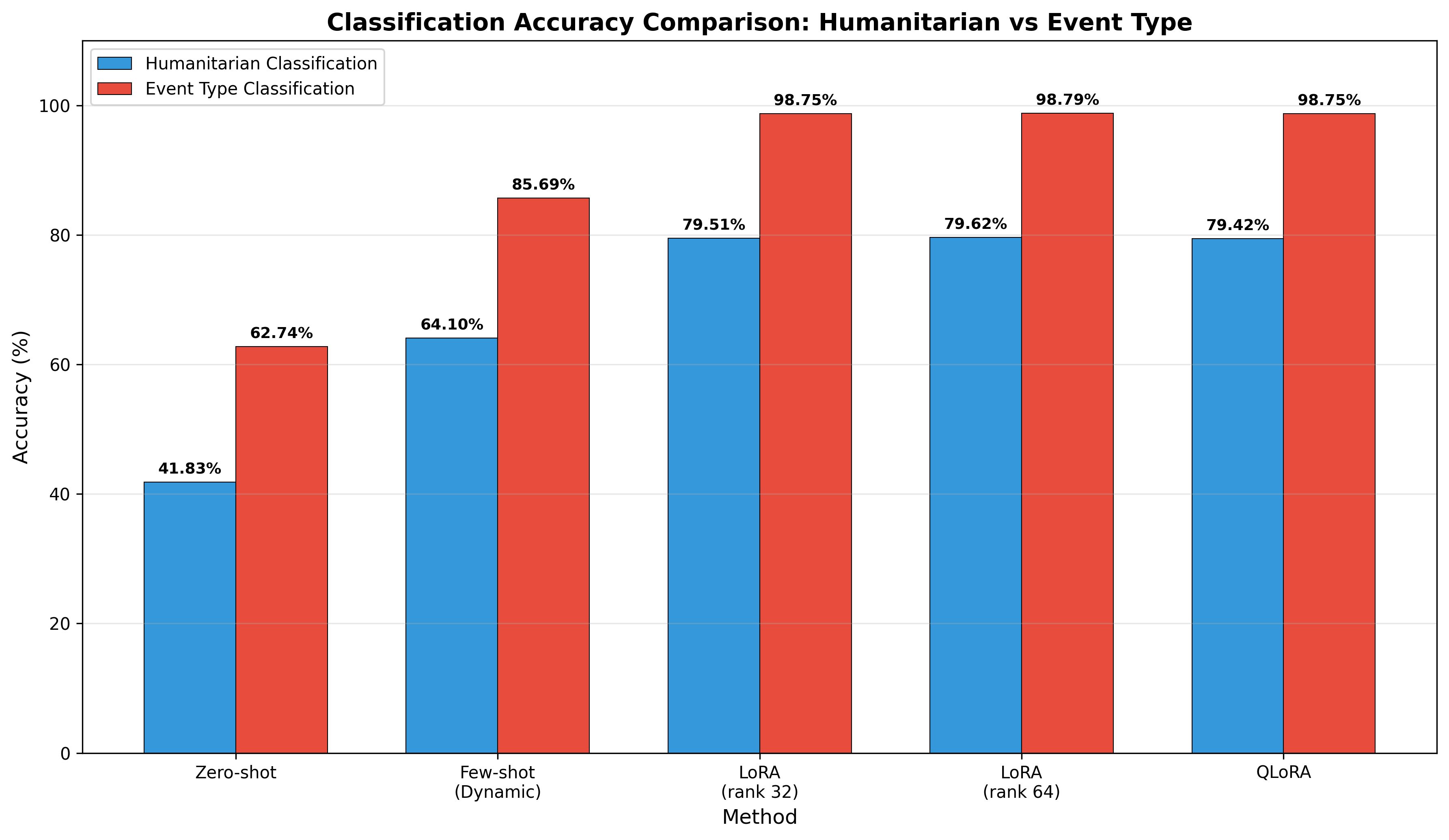}
    \caption{Overall classification accuracy comparison: Humanitarian vs Event Type classification. LoRA fine-tuning achieves approximately 79.6\% accuracy on humanitarian classification and approximately 98.8\% accuracy on event type classification.}
    \label{fig:main_comparison}
\end{figure}

Figure~\ref{fig:main_comparison} summarizes the main experimental results, clearly showing the performance differences across methods for both humanitarian classification and event type classification tasks.

The key findings from this chapter's experiments are:
\begin{enumerate}
    \item \textbf{LoRA is highly effective}: Fine-tuning improves humanitarian classification accuracy from 41.83\% to 79.62\% (+37.79\%) and event type classification accuracy from 62.74\% to 98.79\%
    \item \textbf{RAG shows limited benefit}: Retrieval augmentation not only fails to improve but actually harms classification performance for fine-tuned models
    \item \textbf{Fine-tuned embeddings can backfire}: Task-specific embedding fine-tuning counter-intuitively worsens RAG performance
    \item \textbf{Efficiency matters}: QLoRA achieves 99\% of LoRA performance with 4-bit quantization
\end{enumerate}

These results suggest that for well-defined classification tasks with sufficient training data, \textbf{parameter-efficient fine-tuning (LoRA) is the preferred approach}, while RAG may be better suited for knowledge-intensive tasks where training data is scarce.

\section{Confusion Pair Analysis and Category Ambiguity}
\label{sec:confusion_analysis}

The experimental results presented earlier show that certain humanitarian information categories achieve notably lower classification accuracy, such as ``other\_relevant\_information'' (F1=0.59) and ``not\_humanitarian'' (F1=0.65). This section demonstrates through confusion pair analysis and GPT-4 comparison experiments that these low accuracies are \textbf{not due to model inadequacy}, but rather \textbf{inherent ambiguity in the task itself}—even human annotators or more powerful models struggle to reliably distinguish these categories.

Table~\ref{tab:ambiguous_examples} presents real confusion cases from the test set. These tweets are extremely difficult to classify even for humans, as they simultaneously exhibit semantic features of multiple categories. Consider the first example: the tweet mentions infrastructure damage (``no water, no electricity''), while also expressing community solidarity (``we can go through this together''), yet is labeled as ``not\_humanitarian''. Three classifiers—human annotators, the LoRA model, and GPT-4—produced three different judgments, demonstrating that \textbf{the category boundaries themselves are inherently fuzzy}.

\begin{table}[!t] 
\centering        
\caption{Typical confusion cases difficult even for humans. Each tweet can reasonably be assigned to multiple categories.}
\label{tab:ambiguous_examples}

\resizebox{\columnwidth}{!}{%
\begin{tabular}{p{3.5cm}ccc}
\toprule
\textbf{Tweet Content} & \textbf{Gold Label} & \textbf{Model Pred} & \textbf{GPT-4 Pred} \\
\midrule
``Not water, not electricity. This earthquake has hit us horrible...'' & Not Hum. & Other Rel. & Infra. \\ 

\midrule
``Hurricane Harvey: 70\% of home damage costs aren't covered...'' & Other Rel. & Infra. & Infra. \\
\midrule
``Please consider doing what you can to aid Hurricane Harvey Relief...'' & Sympathy & Rescue & Rescue \\
\midrule
``We urgently need food and water. Red Cross is coming to help.'' & Requests & Rescue & Rescue \\
\bottomrule
\end{tabular}%
}
\end{table}

To assess model inadequacy, GPT-4 reclassified the top 10 confusion pairs (300 samples, Table~\ref{tab:gpt4_analysis}). GPT-4 achieved only 21.67\% accuracy on these errors, notably failing (3.33\%) on the ``other\_relevant\_information'' vs. ``not\_humanitarian'' distinction. Frequency weighting indicates that even full GPT-4 correction would yield only a marginal +2.59\% improvement (79.51\%$\to$82.10\%). Significant error disagreement (22.33\%) between LoRA and GPT-4 confirms that performance is primarily constrained by inherent taxonomy ambiguity rather than model capability.

\begin{table}[!t]
\centering
\caption{GPT-4 performance on confusion samples (category names abbreviated for consistency).}
\label{tab:gpt4_analysis}
\resizebox{\columnwidth}{!}{%
\begin{tabular}{lcccc}
\toprule
\textbf{Confusion Pair} & \textbf{Freq.} & \textbf{GPT-4 Acc.} & \textbf{Corr.} \\
\midrule
Infra. \& Utility $\to$ Other Relevant Info & 118 & 36.67\% & 43 \\
Caution \& Advice $\to$ Other Relevant Info & 197 & 33.33\% & 66 \\
Not Humanitarian $\to$ Other Relevant Info  & 277 & 30.00\% & 83 \\
Requests \& Needs $\to$ Rescue \& Donation  & 130 & 26.67\% & 35 \\
Other Relevant Info $\to$ Caution \& Advice  & 201 & 26.67\% & 54 \\
Other Relevant Info $\to$ Not Humanitarian  & 277 & \textbf{3.33\%} & 9 \\
\midrule
\textbf{Average} & 1,917 & \textbf{21.67\%} & 393 \\
\bottomrule
\end{tabular}%
}
\end{table}

Beyond limited accuracy, GPT-4 also presents significant \textbf{cost and efficiency challenges}: processing 300 samples took approximately 9 minutes and cost around \$3.50, with API calls unable to efficiently parallelize, resulting in inference speeds far slower than local LoRA models. Applying GPT-4 to all 1,917 confusion samples would cost an estimated \$22, improving accuracy by only 2.59\%—\textbf{approximately \$8.50 per percentage point improvement}, an extremely low return on investment.

The confusion matrix in Figure~\ref{fig:lora_confusion} and top confusion pairs analysis in Figure~\ref{fig:lora_confusion} reveal the root of the problem: ``other\_relevant\_information'' is a catch-all category exhibiting bidirectional confusion with almost all other categories; 554 pairs of bidirectional confusion (277+277) exist between ``other\_relevant\_information'' and ``not\_humanitarian'', accounting for 17.8\% of total errors; ``requests\_or\_urgent\_needs'' and ``rescue\_volunteering\_or\_donation\_effort'' share ``help'' semantics—the former asks for help, the latter offers help, but tweets often contain both.

\begin{figure}[!t]  
    \centering

    \includegraphics[width=\linewidth]{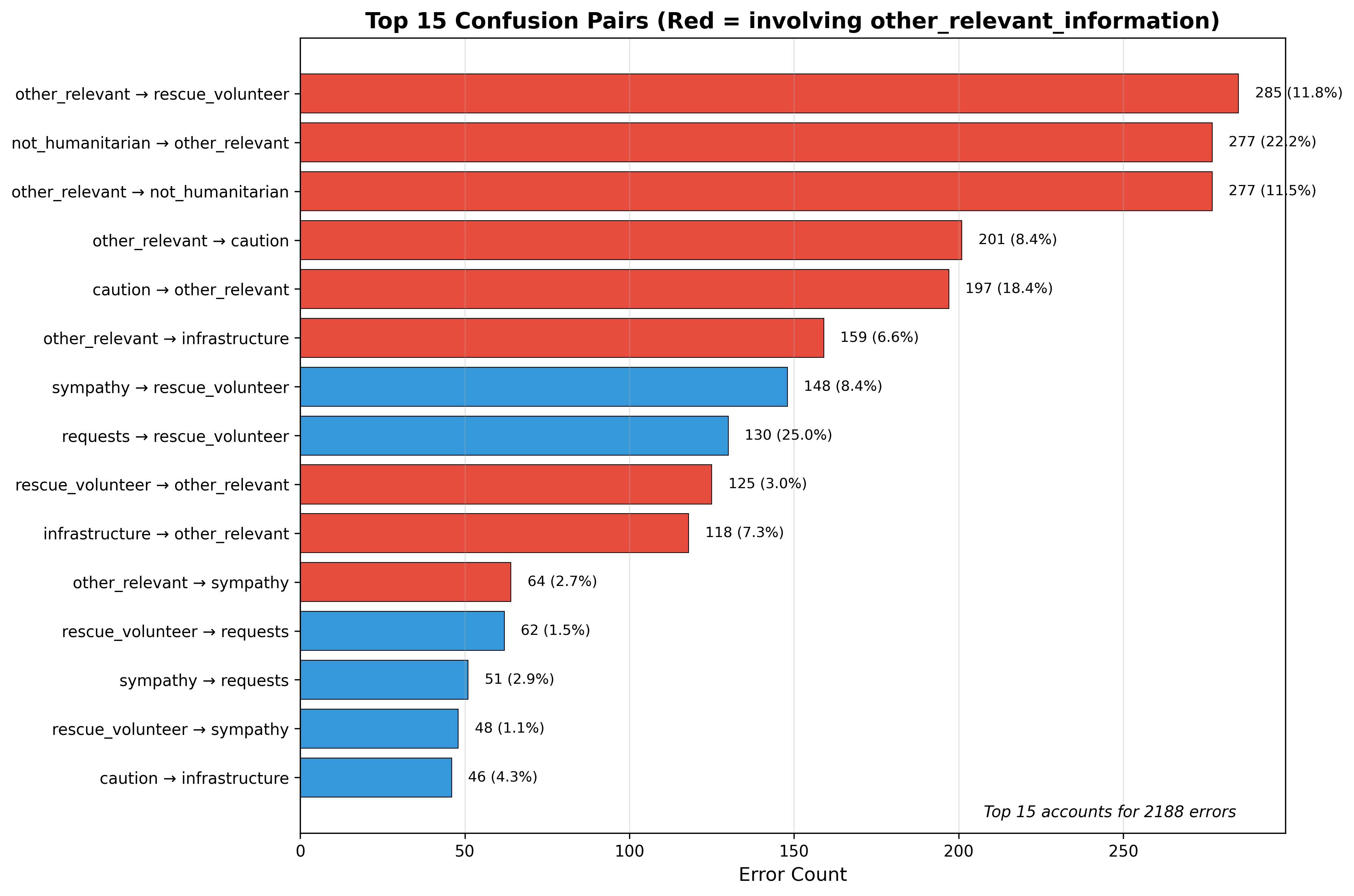}
    \caption{Top 15 confusion pairs analysis. Red bars indicate confusions involving ``other\_relevant\_information'', which exhibits severe bidirectional confusion with almost all other categories.}
    \label{fig:top_confusion_pairs}
\end{figure}

This section's experiments demonstrate that \textbf{the ``low accuracy'' categories in the current model are not due to model inadequacy, but rather inherent ambiguity in the HumAID dataset's category definitions}. The GPT-4 validation experiment confirms: even larger, more expensive models cannot effectively resolve these confusions. Practical recommendations include: (1) accept the ambiguity—for real-world tasks like disaster information classification, some category boundaries are inherently fuzzy; (2) consider merging high-confusion categories into ``general\_information'' to eliminate the largest confusion source; (3) concentrate resources on semantically clear, high-application-value categories. This finding reframes the problem: \textbf{not ``how to make the model better,'' but ``how to design a clearer category system.''}


\section{Conclusion}
\label{sec:conclusion}

This study systematically compared three LLM adaptation paradigms---prompting, parameter-efficient fine-tuning (LoRA), and retrieval-augmented generation (RAG)---for disaster-related humanitarian information classification. 

\paragraph{Key Findings}
Experiments on the HumAID dataset using Llama 3.1 8B yield the following conclusions:
\begin{enumerate}
    \item \textbf{LoRA is the superior adaptation strategy}, improving humanitarian classification accuracy from 41.83\% to 79.62\% (+37.79\%) and event type accuracy to 98.79\%. This underscores the necessity of task-specific fine-tuning for fine-grained classification.
    \item \textbf{RAG utility is inversely correlated with model capability}: while RAG benefits baseline models (+13.14\%), it degrades fine-tuned LoRA models (-0.32\% to -2.09\%), suggesting that retrieved noise can outweigh information in high-performance models.
    \item \textbf{Performance is constrained by taxonomy ambiguity}: persistent confusion between ``other\_relevant\_information'' and ``not\_humanitarian'' reflects inherent label overlap. GPT-4's low accuracy (21.67\%) on these samples confirms a bottleneck in category definitions rather than model architecture.
    \item \textbf{QLoRA enables efficient deployment}, maintaining 99.4\% of LoRA's performance while halving memory requirements, offering a viable solution for resource-constrained disaster response.
\end{enumerate}

\paragraph{Research Contributions}
We provide the first systematic comparison of prompting, LoRA, and RAG for crisis informatics, identifying a critical inverse correlation between model strength and RAG effectiveness. Furthermore, we redefine the task's performance ceiling by identifying inherent label ambiguity as the primary constraint.

\paragraph{Practical Recommendations}
For operational deployment, we recommend: (1) prioritizing LoRA/QLoRA over RAG for classification; (2) merging semantically overlapping categories like ``other\_relevant\_information'' to eliminate ambiguity; and (3) focusing resources on high-value labels such as ``injured\_or\_dead\_people''.

\paragraph{Future Work}
Future research should focus on taxonomy redesign in collaboration with domain experts. Technical directions include cross-disaster transfer learning, multilingual support, and real-time streaming optimization for emergency response.

\paragraph{Concluding Remarks}
This study demonstrates that for well-defined classification tasks, \textbf{parameter-efficient fine-tuning is the optimal choice}, whereas RAG is better suited for knowledge-intensive domains. This framework provides an efficient, reliable, and resource-sensitive approach to automated disaster intelligence, supporting timely humanitarian response.

\bibliographystyle{IEEEtran}

\bibliography{disasterLLM}



\appendix

\input{appendix} 
\end{document}

%% file: appendix.tex
\section{Prompt Templates and Auxiliary Details}
\label{app:prompts}

\subsection{Illustrative Failure Modes of Poor Prompt Design}
\label{app:prompt_failures}

This appendix provides representative examples showing how poorly constrained prompts can yield schema-violating generations (free-form text, invented labels, or verbose explanations). These examples are used for illustration only and are not part of the formal evaluation.

\paragraph{Bad prompt variant 1: Unconstrained prompt}\par\noindent
\begin{tcolorbox}[colback=red!3, colframe=red!40, title=Unconstrained Prompt (Illustration)]
\small
\textbf{Prompt:}\\
\texttt{Classify: \{tweet\}}\\[0.5em]
\textbf{Example output (schema-violating):}\\
\texttt{\#disasterrelief ... (free-form content / hashtags / explanations)}
\end{tcolorbox}

\paragraph{Bad prompt variant 2: Vague instruction}\par\noindent
\begin{tcolorbox}[colback=red!3, colframe=red!40, title=Vague Instruction (Illustration)]
\small
\textbf{Prompt:}\\
\texttt{What category does this disaster tweet belong to?}\\
\texttt{Tweet: \{tweet\}}\\[0.5em]
\textbf{Example output (schema-violating):}\\
\texttt{The tweet belongs to the category of condolences... Explanation: ...}
\end{tcolorbox}

\paragraph{Bad prompt variant 3: Incomplete label set}\par\noindent
\begin{tcolorbox}[colback=red!3, colframe=red!40, title=Incomplete Labels (Illustration)]
\small
\textbf{Prompt:}\\
\texttt{Classify the following tweet into a humanitarian category.}\\
\texttt{Tweet: \{tweet\}}\\[0.5em]
\textbf{Example output (schema-violating):}\\
\texttt{Humanitarian category: Support and Solidarity ... (non-standard label)}
\end{tcolorbox}

\subsection{Label Spaces}
\label{app:label_spaces}

For reproducibility, we list the full label spaces used across all prompting baselines.

\begin{tcolorbox}[colback=gray!4, colframe=gray!45, title=Humanitarian Label Set $\mathcal{Y}_h$ (10 classes)]
\small
\texttt{caution\_and\_advice}\\
\texttt{displaced\_people\_and\_evacuations}\\
\texttt{infrastructure\_and\_utility\_damage}\\
\texttt{injured\_or\_dead\_people}\\
\texttt{missing\_or\_found\_people}\\
\texttt{not\_humanitarian}\\
\texttt{other\_relevant\_information}\\
\texttt{requests\_or\_urgent\_needs}\\
\texttt{rescue\_volunteering\_or\_donation\_effort}\\
\texttt{sympathy\_and\_support}
\end{tcolorbox}

\begin{tcolorbox}[colback=gray!4, colframe=gray!45, title=Event Type Set $\mathcal{Y}_e$ (4 classes)]
\small
\texttt{earthquake, fire, flood, hurricane}
\end{tcolorbox}

\subsection{Zero-shot Prompt Template}
\label{app:zeroshot_prompt}

We use a well-constrained prompt that enforces (i) full label enumeration, (ii) a single JSON object with fixed field names, and (iii) no explanations.

\begin{tcolorbox}[colback=gray!5, colframe=gray!50, title=Zero-Shot Prompt Template (Main Experiments)]
\small
\texttt{You are an expert disaster tweet classifier.}\\
\texttt{You must classify each tweet into TWO fields:}\\[0.4em]
\texttt{1) Humanitarian Label (choose exactly ONE):}\\
\texttt{caution\_and\_advice, displaced\_people\_and\_evacuations, infrastructure\_and\_utility\_damage, injured\_or\_dead\_people, missing\_or\_found\_people, not\_humanitarian, other\_relevant\_information, requests\_or\_urgent\_needs, rescue\_volunteering\_or\_donation\_effort, sympathy\_and\_support}\\[0.4em]
\texttt{2) Event Type (choose exactly ONE):}\\
\texttt{earthquake, fire, flood, hurricane}\\[0.6em]
\texttt{Return ONLY ONE JSON object. No explanation.}\\
\texttt{Use this EXACT format:}\\
\texttt{\{"humanitarian\_label": "...", "event\_type": "..."\}}\\[0.6em]
\texttt{Tweet: \{tweet\}}
\end{tcolorbox}

\subsection{Few-shot Prompt Structure}
\label{app:fewshot_prompt}

Few-shot prompting prepends $k$ labeled demonstrations before the test instance. The same schema constraints are applied.

\begin{tcolorbox}[colback=blue!4, colframe=blue!45, title=Few-Shot Prompt Structure (Chat Style)]
\small
\textbf{System:}\\
\texttt{You are an expert disaster tweet classifier.}\\
\texttt{Classify each tweet into TWO fields and return ONLY ONE JSON object. No explanation.}\\
\texttt{Humanitarian Label (choose 1): [FULL LIST AS IN Section~\ref{app:zeroshot_prompt}]}\\
\texttt{Event Type (choose 1): earthquake, fire, flood, hurricane}\\[0.6em]
\textbf{User (Demo 1):}\\
\texttt{Tweet: ``[demo tweet 1]''}\\
\textbf{Assistant (Demo 1):}\\
\texttt{\{"humanitarian\_label": "...", "event\_type": "..."\}}\\[0.4em]
\textbf{User (Demo 2):}\\
\texttt{Tweet: ``[demo tweet 2]''}\\
\textbf{Assistant (Demo 2):}\\
\texttt{\{"humanitarian\_label": "...", "event\_type": "..."\}}\\[0.4em]
\texttt{... repeat for $k$ demos ...}\\[0.6em]
\textbf{User (Test):}\\
\texttt{Tweet: ``\{tweet\}''}\\
\textbf{Assistant:}\\
\texttt{[MODEL OUTPUT JSON]}
\end{tcolorbox}

\subsection{Few-shot Demonstration Construction}
\label{app:fewshot_examples}

\subsubsection{Static Stratified Sampling}
\label{app:static_sampling}

We build a static set of $k$ demonstrations with stratification over the humanitarian label space $\mathcal{Y}_h$.

\begin{algorithm}[H]
\caption{Stratified Static Example Selection (over $y_h$)}
\label{alg:static_sampling_app}
\begin{algorithmic}[1]
\Require Training set $\mathcal{D}_{train}$, Number of shots $k$, Humanitarian label set $\mathcal{Y}_h$
\Ensure Example set $\mathcal{E}$
\State $\mathcal{E} \gets \emptyset$
\For{each label $l \in \mathcal{Y}_h$}
    \If{$|\mathcal{E}| < k$}
        \State $\mathcal{D}_l \gets \{(x, y_h, y_e) \in \mathcal{D}_{train}: y_h = l\}$
        \If{$|\mathcal{D}_l| > 0$}
            \State Sample $e \sim \text{Uniform}(\mathcal{D}_l)$
            \State $\mathcal{E} \gets \mathcal{E} \cup \{e\}$
        \EndIf
    \EndIf
\EndFor
\While{$|\mathcal{E}| < k$}
    \State Sample $e \sim \text{Uniform}(\mathcal{D}_{train} \setminus \mathcal{E})$
    \State $\mathcal{E} \gets \mathcal{E} \cup \{e\}$
\EndWhile
\State \Return $\mathcal{E}$
\end{algorithmic}
\end{algorithm}

\subsubsection{Naive Retrieval-Augmented Demonstrations (TF-IDF)}
\label{app:tfidf_retrieval}

For the naive retrieval-augmented setup, demonstrations are retrieved from the training pool via TF-IDF cosine similarity. We use the standard TF-IDF weighting:
\begin{equation}
\text{TF-IDF}(t, d, D) = \text{TF}(t, d)\times \log\frac{|D|}{|\{d'\in D: t\in d'\}|}.
\end{equation}

\subsection{Inference Configuration}
\label{app:inference_config}

All prompting baselines are evaluated using deterministic decoding: $\text{temperature} = 0.0$, $\text{top\_p} = 1.0$, $\texttt{max\_tokens} = 50$.

\section{RAG Implementation Details}
\label{app:rag_details}

\subsection{Embedding Fine-tuning Configuration}
\label{app:embedding_config}

Table~\ref{tab:embedding_config_app} summarizes the embedding model fine-tuning configuration.

\begin{table}[H]
\begin{center}
\caption{Embedding model fine-tuning configuration}
\label{tab:embedding_config_app}
\begin{tabular}{ll}
\toprule
\textbf{Parameter} & \textbf{Value} \\
\midrule
Base Model & all-MiniLM-L6-v2 \\
Training Pairs per Sample & 2 (1 positive + 1 negative) \\
Batch Size & 32 \\
Epochs & 2 \\
Warmup Steps & 10\% of total steps \\
Loss Function & CosineSimilarityLoss \\
\bottomrule
\end{tabular}
\end{center}
\end{table}

\subsection{Hybrid Arbitration Algorithm}
\label{app:hybrid_algorithm}

\begin{algorithm}[H]
\caption{Hybrid Arbitration Reranking}
\label{alg:hybrid_arbitration_app}
\begin{algorithmic}[1]
\Require Retrieved neighbors $\mathcal{N}$, Phase-1 prediction $\hat{y}^{(1)}$, top-$k$ count
\Ensure Reranked neighbor indices

\State $\mathcal{S} \gets \{n \in \mathcal{N} : n.\text{label} = \hat{y}^{(1)}\}$ \Comment{Supporters}

\If{$|\mathcal{S}| > 0$}
    \State \Return $\mathcal{S}[0:k]$ \Comment{Prioritize consistent examples}
\EndIf

\State $\text{counts} \gets \text{Counter}([n.\text{label} \text{ for } n \in \mathcal{N}])$
\State $(\text{dominant\_label}, \text{count}) \gets \text{most\_common}(\text{counts})$

\If{$\text{count} \geq 0.5 \times |\mathcal{N}|$}
    \State \Return examples with dominant\_label $[0:k]$ \Comment{Strong correction}
\EndIf

\State \Return original top-$k$ \Comment{Fallback}
\end{algorithmic}
\end{algorithm}

\subsection{RAG Prompt Template}
\label{app:rag_prompt}

\begin{tcolorbox}[colback=gray!5, colframe=gray!40, title=RAG Prompt Template]
\small
\textbf{System}: You are an expert disaster tweet classifier...\\[0.3em]
\textbf{User}: Here are some similar examples for reference:\\[0.2em]
Tweet: ``[retrieved tweet 1]''\\
JSON: \{``humanitarian\_label'': ``[label 1]'', ...\}\\[0.2em]
...\\[0.2em]
Tweet: ``[test tweet]''
\end{tcolorbox}